\documentclass[10pt,twocolumn,letterpaper]{article}

\usepackage[pagenumbers]{cvpr2023/cvpr} % To force page numbers, e.g. for an arXiv version % if put this file in the top folder

\usepackage{graphicx}
\usepackage{caption}
\usepackage{amsmath,amssymb,enumerate}

\usepackage{amsthm}
\usepackage{mathtools}
\usepackage{breqn}
\usepackage{algorithm, algorithmicx, algpseudocode}

\usepackage{blindtext}
\usepackage{gensymb}
\usepackage{xparse}
\usepackage{lipsum}
\usepackage{mathrsfs}
\usepackage[mathscr]{euscript}
\usepackage{times}
\usepackage{cite} % to group references
\usepackage{balance}
\usepackage{comment}

\renewcommand{\thefigure}{\arabic{figure}}

\newcommand{\abs}[1]{\left\lvert#1\right\rvert}

\usepackage{booktabs}
\usepackage{multirow}
\usepackage{stfloats}
\usepackage[accsupp]{axessibility}

\usepackage{hyperref}
\hypersetup{breaklinks,colorlinks}

\usepackage[capitalize]{cleveref}
\crefname{section}{Sec.}{Secs.}
\Crefname{section}{Section}{Sections}
\Crefname{table}{Table}{Tables}
\crefname{table}{Tab.}{Tabs.}

\def\cvprPaperID{10471} % *** Enter the CVPR Paper ID here
\def\confName{CVPR}
\def\confYear{2023}

\begin{document}

\title{E2PN: Efficient SE(3)-Equivariant Point Network}

\author{Minghan Zhu\\
University of Michigan\\
{\tt\small minghanz@umich.edu}
\and
Maani Ghaffari\\
University of Michigan\\
{\tt\small maanigj@umich.edu}
\and 
William A Clark \\
Cornell University \\
{\tt\small wac76@cornell.edu}
\and
Huei Peng\\
University of Michigan\\
{\tt\small hpeng@umich.edu}
}
\maketitle

%%%%%%%%% ABSTRACT
\begin{abstract}
This paper proposes a convolution structure for learning SE(3)-equivariant features from 3D point clouds. It can be viewed as an equivariant version of kernel point convolutions (KPConv), a widely used convolution form to process point cloud data. Compared with existing equivariant networks, our design is simple, lightweight, fast, and easy to be integrated with existing task-specific point cloud learning pipelines. We achieve these desirable properties by combining group convolutions and quotient representations. Specifically, we discretize SO(3) to finite groups for their simplicity while using SO(2) as the stabilizer subgroup to form spherical quotient feature fields to save computations. We also propose a permutation layer to recover SO(3) features from spherical features to preserve the capacity to distinguish rotations. Experiments show that our method achieves comparable or superior performance in various tasks, including object classification, pose estimation, and keypoint-matching, while consuming much less memory and running faster than existing work. The proposed method can foster the development of equivariant models for real-world applications based on point clouds. 
\end{abstract}

%%%%%%%%% BODY TEXT
\section{Introduction}
\label{sec:intro}

Processing 3D data has become a vital task today as demands for automated robots and augmented reality technologies emerge. In the past decade, computer vision has significantly succeeded in image processing, but learning from 3D data such as point clouds is still challenging. An important reason is that 3D data presents more variations than 2D images in several aspects. For example, the rigid body transformations in 2D only have 3 degrees of freedom (DoF) with 1 for rotations. In 3D space, the DoF is 6, with 3 for rotations. 
The 
% sliding-window mechanism enabling the 
2D translation equivariance is a key factor in the success of convolutional neural networks (CNNs) in image processing,
% because it enables parameter sharing and generalization across 2D translations. 
but it is not enough for 3D tasks.

Generally speaking, \textit{equivariance} is a property for a map such that given a transformation in the input, the output changes in a predictable way determined by the input transformation. It drastically improves generalization as the variance caused by the transformations is captured via the network by design. Take CNNs as an example, the equivariance property refers to the fact that a translation in the input image results in the same translation in the feature map output from a convolution layer. However, conventional convolutions are not equivariant to rotations, which becomes problematic, especially when we deal with 3D data where many rotational variations occur. In response, on the one hand, data augmentations with 3D rotations are frequently used. On the other hand, equivariant feature learning emerges as a research area, aiming to generalize the translational equivariance to broader transformations. %We will mainly discuss equivariant feature learning, to which our work also belongs. 

\begin{figure}
  \centering
  \includegraphics[width=\linewidth]{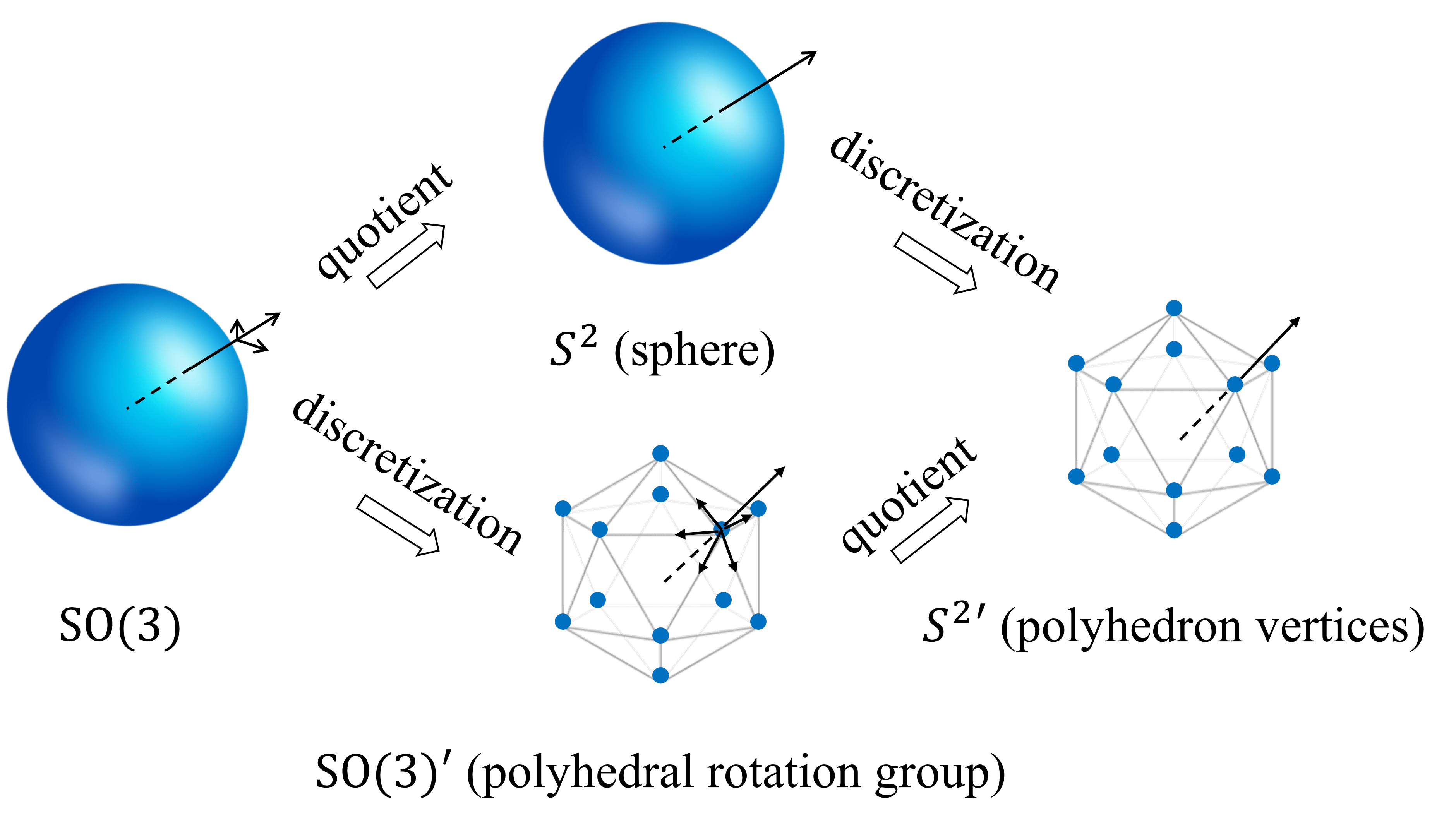}
  \caption{Our method achieves higher efficiency by working with smaller feature maps defined on $S^2{'}\times \mathbb{R}^3$ rather than $SO(3)'\times\mathbb{R}^3$ ($'$ denotes discretization). $\mathbb{R}^3$ is omitted in the figure. The black arrows in each space represent elements. The top and bottom paths are equivalent, showing the relations among different representations. }
  \label{fig:intro}
\end{figure}

A lot of progress has been made in group-equivariant feature learning. The term \textit{group} encompasses the 3D rotations and translations, which is called the special Euclidean group of dimension 3, denoted $\mathrm{SE}(3)$, and also other more general types of transformations that represent certain symmetries. 
% Technical details will be discussed later. 
While many methods have been proposed,
%to learn features equivariant to various transformations,
% , including but not limited to 2D rotations and translations, 3D rotations and translations, permutations, and higher-order transformations which may not have an intuitive geometric meaning. However, 
equivariant feature learning has not yet become the default strategy for 3D deep learning tasks. From our understanding, two major reasons hinder the broader application of equivariant methods. First, networks dealing with continuous groups typically require specially designed operations not commonly used in neural networks, such as generalized Fourier transform and Monte Carlo sampling. Thus, incorporating them into general neural networks for 3D learning tasks is challenging. Second, for the strategy of working on discretized (finite) groups~\cite{cohen2016group,hoogeboom2018hexaconv}, while the network structures are simpler and closer to conventional networks, they usually suffer from the high dimensionality of the feature maps and convolutions, which causes much larger memory usage and computational load, limiting their practical use. 

This work proposes E2PN, a convolution structure for processing 3D point clouds. Our proposed approach can enable $\mathrm{SE}(3)$-equivariance on any network with the KPConv~\cite{thomas2019KPConv} backbone by swapping KPConv with E2PN. The equivariance is up to a discretization on $\mathrm{SO}(3)$. We leverage a quotient representation to save computational and memory costs by reducing $\mathrm{SE}(3)$ feature maps to feature maps defined on $S^2\times \mathbb{R}^3$ (where $S^2$ stands for the 2-sphere). %, which is one fewer degree of freedom. 
Nevertheless, we can recover the full 6 DoF information through a final permutation layer. As a result, our proposed network is $\mathrm{SE}(3)$-equivariant and computationally efficient, ready for practical point-cloud learning applications. 

% Our proposed structure, shown in Fig.~\ref{fig:title}, works on a discretization of $\mathrm{SE}(3)$ so that it is more compatible with conventional neural networks. Meanwhile, we resolve the problem of high dimensionality by defining feature maps not on the (discretized) $\mathrm{SE}(3)$ space but on a quotient space of $\mathrm{SE}(3)$, the $S^2\times \mathbb{R}^3$ space ($\times$ for Cartesian product)\rev{, which is of lower dimension}. Furthermore, we improve the efficiency by designing the convolutional kernels to be symmetric to the discretized rotations. We also propose a permutation layer to recover $\mathrm{SE}(3)$ information from the \rev{lower-dimensional} quotient space when poses are wanted. As a result, our proposed network is both $\mathrm{SE}(3)$-equivariant (approximately up to the discretization) and efficient, ready for application in practical point-cloud learning tasks. 
Overall, this work has the following contributions:
\begin{itemize}
    \item We propose an efficient $\mathrm{SE}(3)$-equivariant convolution structure for 3D point clouds. 
    \item We design a permutation layer to recover the full $\mathrm{SE}(3)$ information from its quotient space. 
    \item We achieve comparable or better performance with significantly reduced computational cost than existing equivariant models.
    \item Our implementation is open-sourced at \url{https://github.com/minghanz/E2PN}. %(url will be provided upon receiving the final decision). 
    % \item Our code will be open-sourced at \url{https://github.com/minghanz/E2PN}. 
    % \item \mgj{How about an open source implementation of the proposed method? Code release is always a major contribution.}
\end{itemize}

% In this paper, we will refrain from referring too much to theoretical formulations, but familiarity with group theory and representation theory is desired. 
Readers can find preliminary introductions to some related background concepts in the appendix.

\section{Related work}
\textbf{Group convolutions:} In 2016, Cohen and Welling proposed G-CNN \cite{cohen2016group}, enabling equivariance beyond translations on 2D images with generalized G-convolutions (group convolutions) over the group of 90-degree rotations, which is one of the earliest efforts in equivariant deep learning. Group convolution is similar to conventional convolutions but has an extended domain for feature maps and kernels. The idea was then applied to different networks to enable equivariance for $\mathrm{SE}(2)$ \cite{cohen2016group, hoogeboom2018hexaconv}, $\mathrm{SO}(3)$ \cite{cohen2019gauge}, $\mathrm{SE}(3)$ \cite{worrall2018cubenet, chen2021equivariant}, and $\mathrm{E}(3)$ \cite{winkels20183d} groups up to some discretization. The idea mainly works with finite (discretized) groups, as it is convenient to parameterize feature maps and kernels on the discretized group elements just as on pixel grids. Group convolutions have a relatively simple structure, making them straightforward to apply, but a major downside is that the lifted domain of features and kernels causes higher computational and memory costs, and the problem is more prominent when the group is large. 
% For example, EPN [] built a SE(3)-equivariance network discretizing SO(3) to the icosahedral rotation group $\mathcal{I}$ with 60 elements, thus lifting the feature domain from $\mathbb{R}^3$ to $\mathbb{R}^3\times \mathcal{I}$, which is a 60-time bump-up. 
Groups convs can also work with continuous groups, for example, with the help of Monte Carlo (MC) estimation as in \cite{finzi2020generalizing}, but they suffer from a large memory burden in MC sampling when the number of layers grows \cite{macdonald2022enabling}. 

\textbf{Steerable CNNs:} Another line of work is steerable CNNs. Instead of augmenting the domain of feature maps, steerable CNNs generalize the space of feature values to be \textit{steerable}, i.e., the feature values transform predictably as the input transforms. The way the feature transforms is called the group representation in the feature space, governed by the feature \textit{type}. Features of \textit{scalar} type are kept unchanged under group actions, and we call the group representation trivial. For \textit{vector} or \textit{tensor} feature types, the representation is not trivial, and the features will change with group actions, for example, through rotation matrix multiplication. Steerable CNNs work for both discretized and continuous groups. One may freely design the feature types based on pre-determined basic types and corresponding representations (irreducible representations, i.e., \textit{irreps}) as building blocks for a given group. Group convolutions can be viewed as a special case of steerable CNNs with \textit{regular} representations, i.e., channels of a feature vector undergo permutations when transformed. Examples of steerable CNNs include \cite{cohen2016steerable, cohen2017convolutional, worrall2017harmonic, weiler20183d, thomas2018tensor}. While the framework of steerable CNNs generalizes group convolutions with more flexibility, it requires a good understanding of representation theory and involves generalized Fourier analysis when working with a continuous group, which makes the structure complicated and challenging to apply in real-world applications broadly. The work of~\cite{weiler2019general, hutchinson2021lietransformer} also found empirically that steerable CNNs with irreps underperform regular representations in certain tasks. 

\textbf{Theoretical progress:} There has been a lot of progress in the theoretical development of equivariant networks that are not group-specific. \cite{kondor2018generalization} developed a general formulation for steerable CNNs with scalar type features, and \cite{cohen2019general, cohen2018intertwiners} generalized it to arbitrary feature types. It is proven that all equivariant linear maps can be written as convolutions. The formulation from the perspective of Fourier analysis was presented in \cite{xu2022unified}. \cite{weiler2019general} found the general form of solution for the equivariant kernels for $E(2)$ group, later generalized to any compact group \cite{lang2020wigner}. \cite{finzi2021practical} proposed an algorithm to solve for the equivariance constraint for arbitrary matrix group. The formulation of group convolution based on MC estimation was proposed for any Lie group with \cite{finzi2020generalizing} or without \cite{macdonald2022enabling} surjective exponential maps. Equivariant non-linear layers like transformers \cite{hutchinson2021lietransformer, fuchs2020se} and equivariant set and graph networks \cite{finzi2021practical, segol2019universal, keriven2019universal} are also proposed, but they are not the focus of this paper. 

% \textbf{Equivariant learning in (large-scale? robotic?) practical perception tasks: }\mhz{I am not sure about the title here, because the molecular or physics tasks can also be considered \textit{practical}, but not practical in the sense we or a lot of audience think. I am not sure whether it will annoy those people if we brand this section of work only as practical. } 
\textbf{Applications of equivariant learning in perception tasks:} We want to highlight a few equivariant networks that gain attention in perception applications due to their simplicity and practicality. Vector Neurons \cite{deng2021vector} is a PointNet-like $\mathrm{SO}(3)$-equivariant network for 3D point cloud learning, later applied to point cloud registration \cite{zhu2022correspondence} and manipulation \cite{simeonov2022neural}. It can be viewed as a special case of TFN \cite{thomas2018tensor} with type-1 features and self-interactions only. E2CNN \cite{weiler2019general} as an $\mathrm{SE}(2)$-equivariant framework was applied in several image processing tasks \cite{Seo_2022_CVPR, Lee_2022_CVPR}, given its generality and user-friendly library. DEVIANT \cite{kumar2022deviant} applied scale-equivariant convolutions in monocular 3D object detection. EPN \cite{chen2021equivariant} is a group convolution network with $\mathrm{SE}(3)$-equivariance for 3D point cloud learning based on KPConv \cite{thomas2019KPConv} and was used in practice, for example in place recognition task \cite{linepn}. We also position this proposed work in this category, aiming to promote the application of equivariant learning with our efficient and easy-to-use design. Our work is developed based on EPN, which also serves as a major baseline in this paper. 
% The theoretical foundation of our method was established in []. 

\section{Methodology}
\subsection{Overview of the idea}
% Here we explain two major things without going into implementation details. First, why we claim our method efficient. Second, how we do not sacrifice expressiveness given the gained efficiency.  
We first explain why our proposed method is more efficient. Then we discuss how we gain efficiency without sacrificing expressiveness. 

\subsubsection{Improved efficiency with quotient features}\label{sec:quo}
Following the idea of group convolutions, one needs to extend the domain of feature maps from the Euclidean space to the group space, from $\mathbb{R}^3$ to $\mathrm{SE}(3)\cong \mathrm{SO}(3)\times \mathbb{R}^3$ in our case, where we want $\mathrm{SE}(3)$-equivariant features for 3D point clouds. We then discretize $\mathrm{SO}(3)$ to a finite group denoted as $\mathrm{SO}(3)'$ (we use $'$ to denote discretization in this paper). The discretization of $\mathbb{R}^3$ is taken care of by KPConv, which is the non-equivariant counterpart of our method, thus, not discussed here. Up to now, we can obtain a $\mathrm{SE}(3)$-equivariant version of KPConv \cite{thomas2019KPConv} realized through group convolution, which is what EPN~\cite{chen2021equivariant} presents. 

The specific form of the finite group $\mathrm{SO}(3)'$ is introduced later in \cref{sec:spec}, but it can be geometrically understood as the set of all rotations that keep a Platonic solid (i.e., a convex and regular 3D polyhedron) unchanged. As illustrated at the bottom of \cref{fig:intro}, the set of rotations can be enumerated by counting the number of vertices, representing rotations taking a given vertex to any vertices, multiplied by the number of edges connected to a single vertex, representing the rotations that keep a vertex fixed. The same result can be achieved by counting the faces or the edges, but we stick with vertices in the following discussion.

In comparison, we propose to define feature maps on $S^2\times \mathbb{R}^3$, where $S^2$ is the sphere space. We have $S^2 = \mathrm{SO}(3)/\mathrm{SO}(2)$, meaning that it is the \textit{quotient space} of $\mathrm{SO}(3)$ given the stabilizer subgroup $\mathrm{SO}(2)$. To understand the quotient space intuitively, we can see that all rotations can be grouped by the destination of a point on the sphere (e.g., the north pole) after the rotation. All rotations bringing the north pole to the same destination point form a \textit{coset}. They are related by rotations around the axis passing through that point, forming a subgroup isomorphic to $\mathrm{SO}(2)$. Thus $S^2$ is the quotient of $\mathrm{SO}(3)$ and $\mathrm{SO}(2)$. In the discretized setup, as depicted in \cref{fig:intro}, $\mathrm{SO}(2)'$ is discretized by the number of edges connected to a single vertex. 
% In the example depicted in \cref{fig:intro}, $\mathrm{SO}(2)'$ has 5 elements of 72$^\circ$ rotations. 
The quotient $S^2{'} = \mathrm{SO}(3)' / \mathrm{SO}(2)'$ corresponds to the vertices on the Platonic solid. 

In general, $|S^2{'}|=|\mathrm{SO}(3)'|/|\mathrm{SO}(2)'| < |\mathrm{SO}(3)'|$, 
% \mhz{I'm not sure whether $\ll$ here is exaggerated. }
thus the size of feature maps and kernels defined on $S^2{'}\times \mathbb{R}^3$ is much smaller than those on $\mathrm{SO}(3)' \times \mathbb{R}^3$. The convolution operation on the former space also requires smaller costs. It is the major reason our method is much more efficient than EPN. There is another design, called \textit{symmetric kernel}, which further improves the efficiency of our method, but we will refer to \cref{sec:sym_ker} for details. 

\begin{figure}
  \centering
  \includegraphics[width=\linewidth]{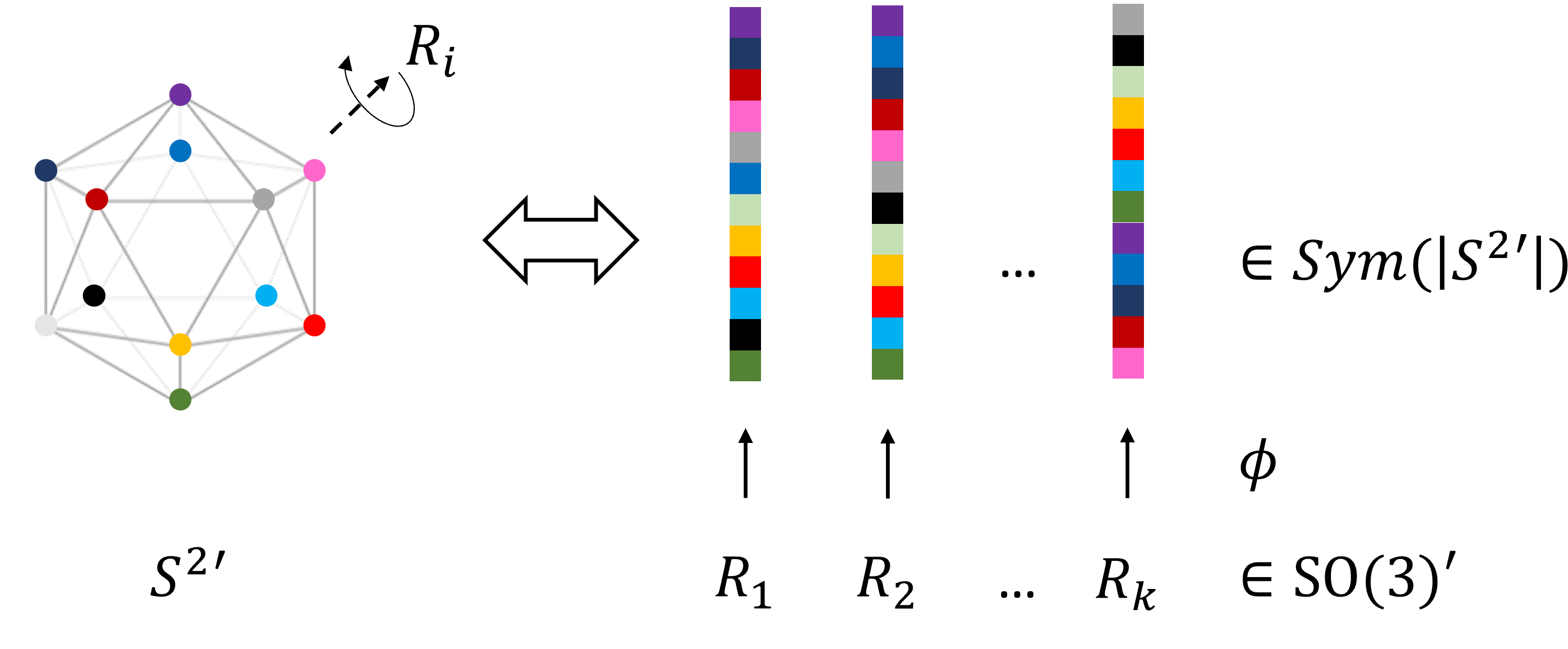}
  \caption{Illustration of recovering $\mathrm{SO}(3)'$ features from $S^2{'}$ features through permutations. The left and right subplots are from the geometric and algebraic views. See \cref{sec:permutation} for more explanation. }
  \label{fig:permutation}
\end{figure}

\subsubsection{Information recovery with permutations}\label{sec:permutation}
With the quotient feature map, all rotations moving the north pole to the same point on a sphere are represented by the same point. Thus we immediately lose the ability to distinguish among these rotations, which is a problem if our task is to learn the pose, for example, in the point cloud registration tasks. 

However, with our proposed permutation layer, we can distinguish every element in $\mathrm{SO}(3)'$ from the feature maps on $S^2{'}$. The key observation is that the action of $\mathrm{SO}(3)$ on $S^2$ is \textit{faithful}, i.e., $\forall R \in \mathrm{SO}(3)$ and $R\neq I$, $\exists x\in S^2$, s.t. $Rx\neq x$. It means that an action of $\mathrm{SO}(3)$ other than identity will always cause a change on the $S^2$ feature map when looking at all points on $S^2$ simultaneously. In the discretized setup, it means that there is an injective map $\phi: \mathrm{SO}(3)' \rightarrow Sym(|S^2{'}|)$, where $Sym(|S^2{'}|)$ stands for the symmetric group, i.e., the collection of all permutations of a set of size $|S^2{'}|$. An element in $Sym(n)$ is a bijective map $\{1,...,n\}\rightarrow \{1,...,n\}$, permuting the indices. In other words, each rotation corresponds to a unique permutation of the $S^2{'}$ elements, from which we can distinguish each rotation from the feature map defined on $S^2{'}$.

% As depicted in \cref{fig:permutation}, by stacking the features on $S^2{'}$ elements according to the distinctive permutations given by $\phi$, we obtain a feature map defined on $\mathrm{SO}(3)'$, so that pose-related tasks can be conducted. 

Here we provide the specific form of the permutation layer. Given $\phi$ mentioned above and a feature map $f: S^2{'} \rightarrow \mathbb{R}^m$, we can build a feature map $\tilde{f}: \mathrm{SO}(3)' \rightarrow \mathbb{R}^{mn}$ defined as: 
\begin{equation}\label{eq:permute}
\tilde{f}(R) = [f(x[\phi_R(1)]), f(x[\phi_R(2)]), ..., f(x[\phi_R(n)]) ]
\end{equation}
where $R\in \mathrm{SO}(3)'$, $n=|S^2{'}|$, and $x[i]\in S^2{'}$ for $i=1, ..., n$. Given that $\phi$ is injective, $\tilde{f}(R_1)\neq \tilde{f}(R_2)$ when $R_1\neq R_2$. Thus we can distinguish $\mathrm{SO}(3)'$ rotations from $\tilde{f}$. See \cref{fig:permutation} for an illustration. 
% \mgj{Can you elaborate and maybe expand this paragraph? I think readers who are not familiar won't learn anything. More intuitive explanations and educational writing will be great here.}

\subsection{Specific form of convolution}\label{sec:spec}
\subsubsection{Recap of KPConv}\label{sec:kpconv}
\begin{figure}
  \centering
  \includegraphics[width=\linewidth]{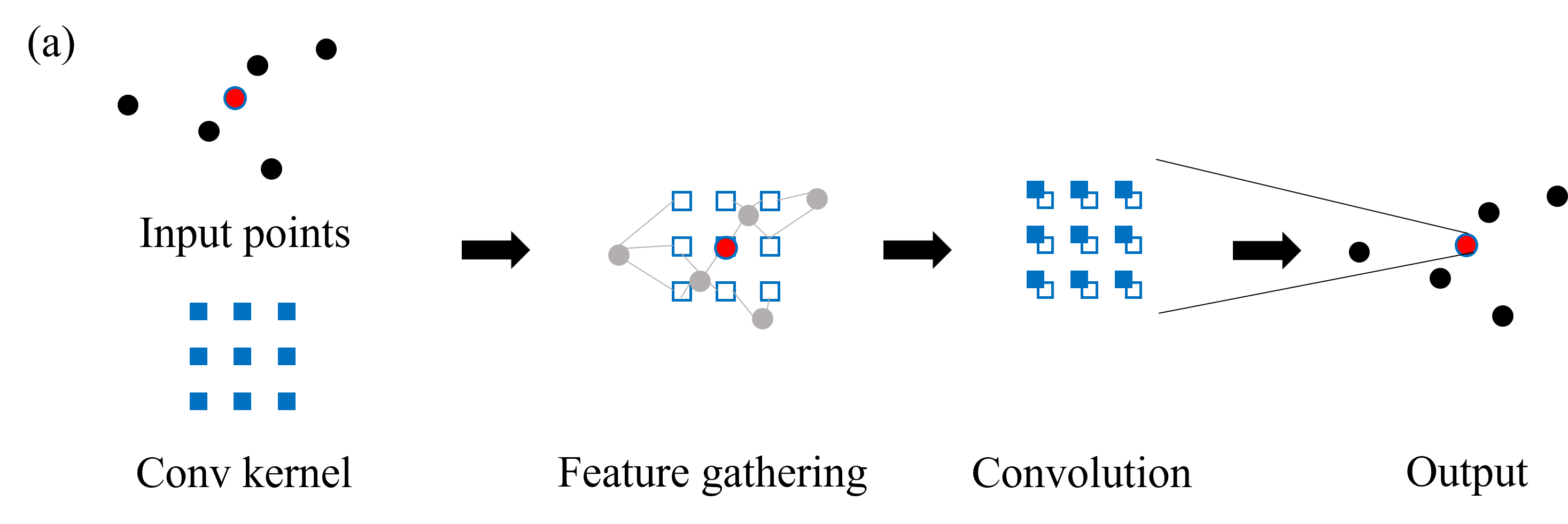}
  \includegraphics[width=\linewidth]{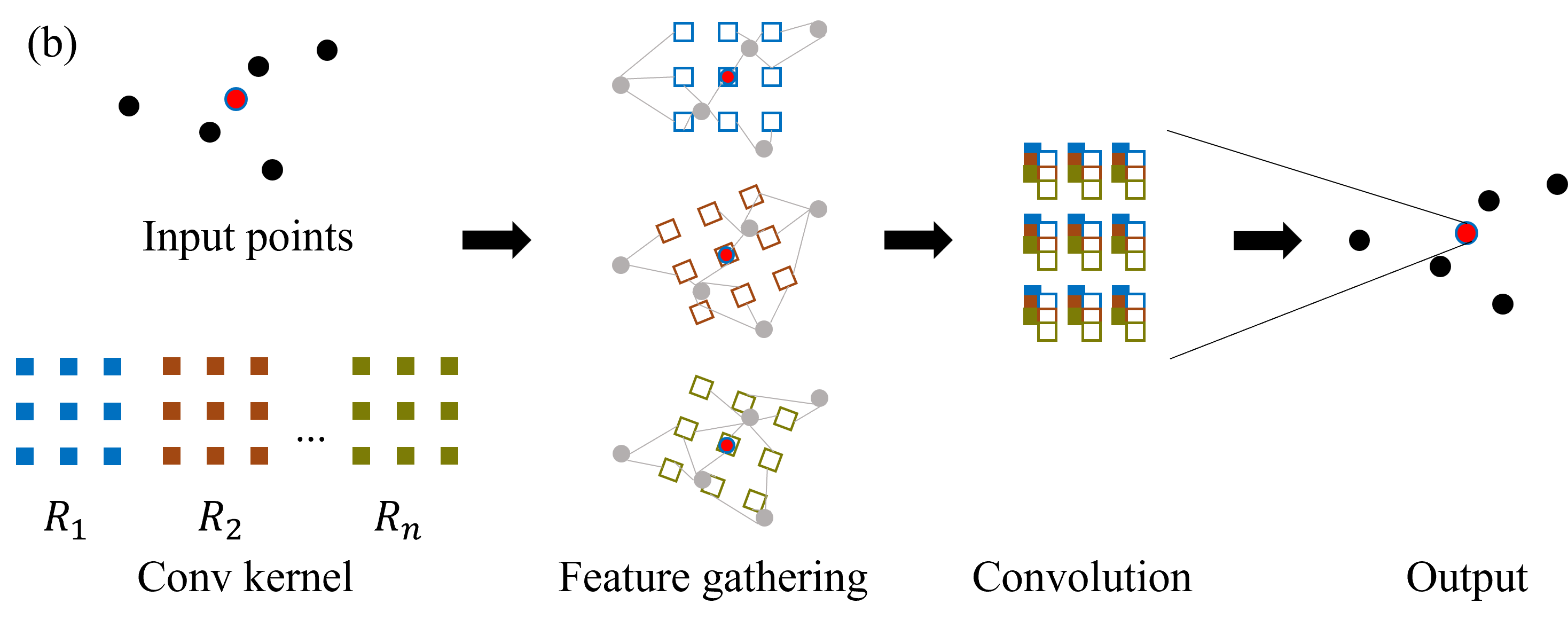}
  \includegraphics[width=\linewidth]{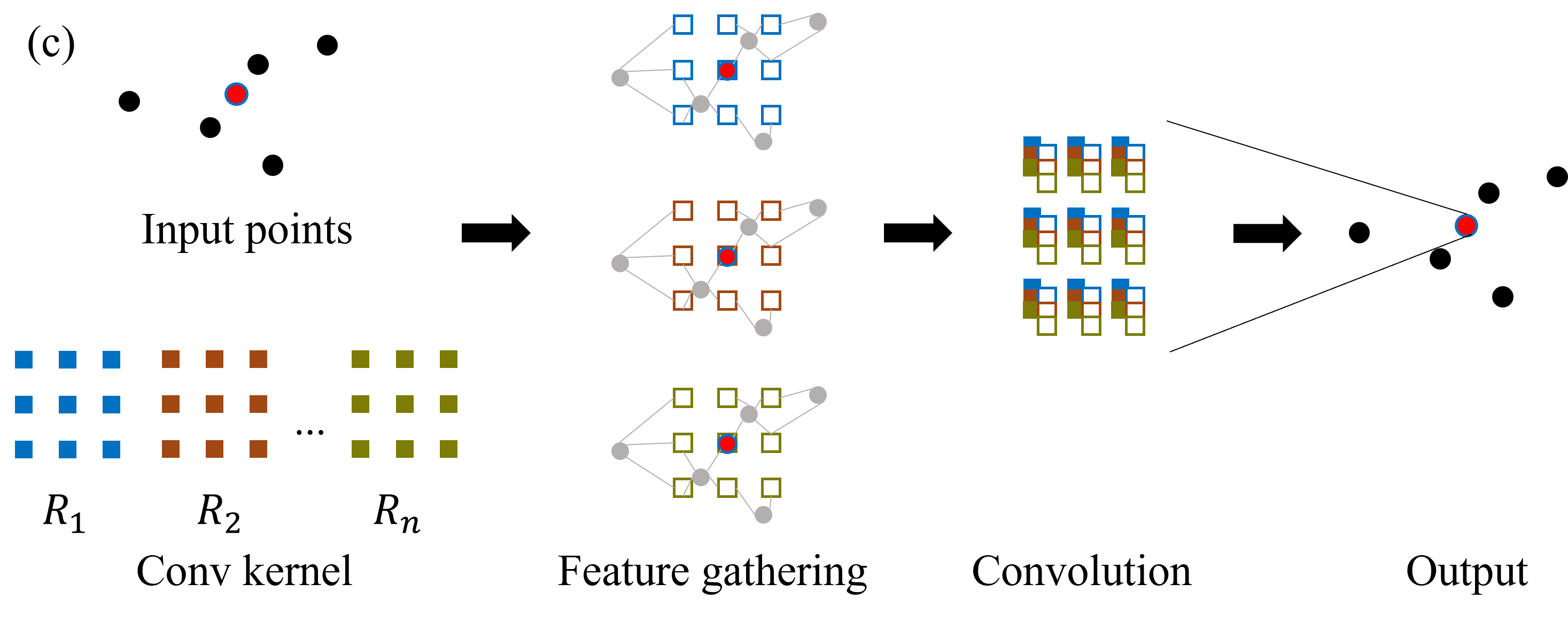}
  \caption{Illustration of different versions of KPConv. (a) original KPConv (see \cref{sec:kpconv}). (b) KPConv on finite group or quotient space without the symmetric kernel (see \cref{sec:kpgroup}). (c) KPConv on group or quotient space with the symmetric kernel (see \cref{sec:sym_ker}). (b,c) compared with (a): the convolution kernel is larger because it is defined on a higher dimension. (c) compared with (b): the feature gathering is more efficient because the kernel points are symmetric to the rotations in the rotation group.
  % \mgj{Add more explanation to keep the figure independent. It is only later, after reading the text it becomes clear what's the difference.}
  }
  \label{fig:sym_kernel}
\end{figure}

A conventional 3D convolution can be written as:
\begin{equation}\label{eq:conv}
    [\kappa * f](x) = \int_{\mathbb{R}^3} \kappa(t)f(x+t)\text{d}t = \sum_{t\in \mathbb{R}^3{'}} \kappa(t)f(x+t),
\end{equation}
where $x\in \mathbb{R}^3$, and the right hand side is after discretization. We use \textit{correlations} to implement convolutions in this paper following conventions in deep learning. For processing point clouds, \cref{eq:conv} could be tricky in implementation because it is challenging to align $\kappa$ with $f$ when there is no grid. The strategy of KPConv is to have a set of kernel points for $\kappa$ and to gather features to the kernel point coordinates from the input points where $f$ is defined so that they are aligned before the convolution. It is depicted in \cref{fig:sym_kernel}~(a), and we need to replace $f$ with $\hat{f}$ in \cref{eq:conv} where $\hat{f}(x)=\sum_{y\in \mathcal{N}_x}w(|y-x|)f(y)$ and $w$ is a scalar weight function based on distance. 
% \mgj{$w(y-x)$ is not defined/explained anywhere. Is it a scalar weight function?} 
$\hat{f}$ is the features gathered from neighboring input points to align with the kernel points. In this case, the input and output feature map $f$ and $[\kappa*f]$ are defined on $\mathbb{R}^3$, while the kernel $\kappa$ is defined on $\mathbb{R}^3{'}$, i.e., coordinates of the set of kernel points. The gathered feature $\hat{f}$ when calculating the convolution at $x\in \mathbb{R}^3$ is defined at $x+\mathbb{R}^3{'}$. Notice that we do not consider the deformable mode of KPConv in this paper. 

\subsubsection{KPConv on group and on quotient space}\label{sec:kpgroup}
To extend KPConv to a group convolution \cite{cohen2016group} with finite group $G'$, we modify \cref{eq:conv} to
\begin{equation}\label{eq:gconv}
    [\kappa * f](g) = \sum_{g_t\in G'} \kappa(g_t)\hat{f}(g\cdot g_t),
\end{equation}
where in our case, $G'=\mathrm{SE}(3)'=\mathrm{SO}(3)'\times \mathbb{R}^3{'}$ is where $\kappa$ is defined, $g\in G^\dagger=\mathrm{SO}(3)'\times \mathbb{R}^3$ is where the input and output feature map is defined. $\hat{f}$ is defined on $g\cdot G'$ when calculating the convolution at $g$. Denote an element $g\in \mathrm{SE}(3)$ as $(R, t)$ where $R\in \mathrm{SO}(3)$ and $t\in \mathbb{R}^3$, the binary operation $\cdot$ for $\mathrm{SE}(3)$ can be specified as $(R_1, t_1)\cdot (R_2, t_2) = (R_1 R_2, t_1 + R_1 t_2)$, same for the discretized case. EPN \cite{chen2021equivariant} follows this form of convolution. However, as discussed in \cref{sec:quo}, the size of group $|\mathrm{SE}(3)'|=|\mathrm{SO}(3)'||\mathbb{R}^3{'}|$, making it expensive to store the feature map and compute the convolution. 
To alleviate this issue, EPN has to conduct convolutions in $\mathrm{SO}(3)'$ and $\mathbb{R}^3{'}$ separately (i.e., separable convolutions \cite{chen2021equivariant, chollet2017xception}), so that the computational cost is reduced.

We use quotient feature maps to address this issue. To conduct convolutions on the quotient space $X=\mathrm{SE}(3)/\mathrm{SO}(2)=S^2\times \mathbb{R}^3$, we cannot directly use \cref{eq:gconv}, because the $\cdot$ operation is not defined between two elements in $X$ since $X$ is not a group. It is pointed out in \cite{cohen2018intertwiners, cohen2019general} that we can write convolutions on the quotient space as 
\begin{equation}\label{eq:qconv}
    [\kappa * f](x) = \sum_{x_t\in X'} \kappa(x_t)\hat{f}(s(x)\cdot x_t),
\end{equation}
where $X'=S^2{'}\times \mathbb{R}^3{'}$ is the domain of $\kappa$, $X^\dagger = S^2{'}\times \mathbb{R}^3$ is the domain of $f$ and $[\kappa*f]$, and $\hat{f}$ is defined on $s(x)\cdot X'$. $s(x)$ is called the \textit{section} map, $s: X'\rightarrow G'$ (or $X\rightarrow G$ in the continuous case), mapping $x\in X'$ to an element of $G'$ in the corresponding coset. With the section map, the $\cdot$ operation denotes the action of the group $G$ on quotient space $X$, i.e., $\cdot: G\times X \rightarrow X$. Denote an element in $S^2\times \mathbb{R}^3$ as $(R\mathbf{n}, t)$, where $R\in \mathrm{SO}(3), \mathbf{n}$ is the north pole point on the unit sphere $(0,0,1)$. Then $R\mathbf{n}$ represents arbitrary points on the sphere $S^2$. The action $\cdot$ can then be written as $(R_1, t_1) \cdot (R_2\mathbf{n}, t_2) = (R_1R_2\mathbf{n}, t_1 + R_1 t_2)$. 

\subsubsection{Symmetric kernels}\label{sec:sym_ker}
% steerability constraint and efficiency
% A question to be answered now is
Now we introduce 
the specific form of $\mathbb{R}^3{'}$, i.e., the location of kernel points in KPConv. In our design, $\mathbb{R}^3{'} = \{rS^2{'}\cup 0\mid r > 0\}$ 
% \mgj{Do you mean $\{rS^2{'} \cup 0\}$? Just a zero after the bar sign for the set condition is weird.}
, i.e., the set of vertices of the Platonic solid with radius $r$ and the origin point. $\mathbb{R}^3{'}$ is very similar to $S^2{'}$, because we want to make the kernel symmetric to $\mathrm{SO}(3)'$. More precisely, we desire $\mathbb{R}^3{'}$ to be closed under the action of $\mathrm{SO}(3)'$. There are two reasons as follows. 

\begin{figure}
  \centering
  \includegraphics[width=0.9\linewidth]{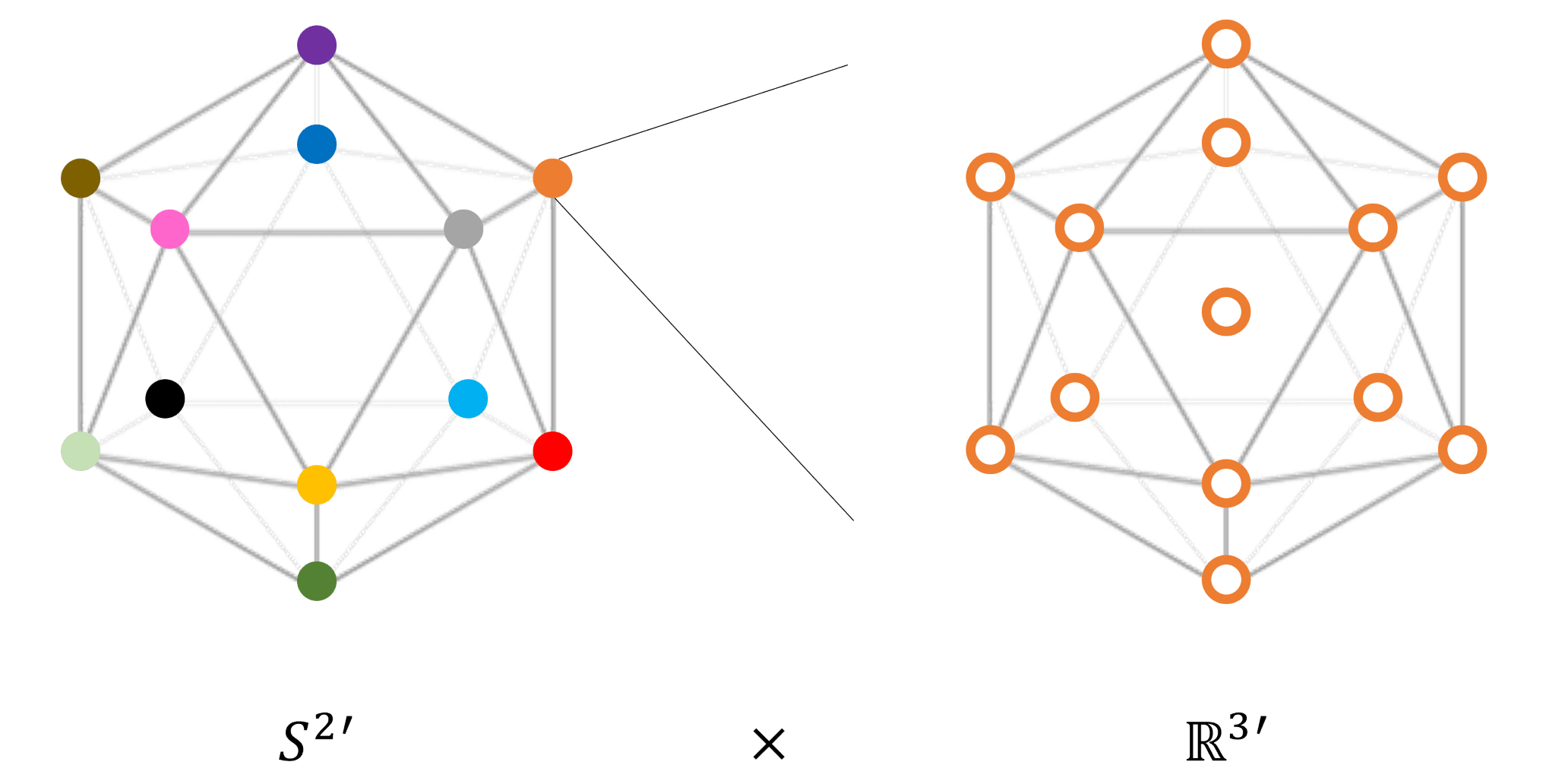}
  \caption{Visualization of the symmetric kernel $\kappa$ used in our work, where $\mathbb{R}^3{'} = \{rS^2{'}\cup 0\mid r > 0\}$. 
  % \mgj{Expand the caption to know what we see. Right now it completely depends on the main text.}
  }
  \label{fig:kernel_spec}
\end{figure}

\paragraph{Steerability constraint}
To make the convolution on quotient space equivariant, defining a valid form of convolution as \cref{eq:qconv} is not the whole  story. The kernel values must satisfy a condition called the \textit{steerability constraint}, which is required for all steerable CNNs. More background knowledge can be found in \cite{cohen2019general}. In our case, the steerability constraint is 
\begin{equation}\label{eq:steerability}
    \kappa(x) = \kappa(R_z\cdot x), \forall x\in X', \forall R_z\in \mathrm{SO}(2)',
\end{equation}
where $R_z$ is a $z$-axis rotation. The derivation of \cref{eq:steerability} from the general form of steerability constraints can be found in the appendix. Here the $\cdot$ operation inherits from the action of $G'$ on $X'$ since $\mathrm{SO}(2)'\subset \mathrm{SO}(3)'\subset G'$. Specifically, we have $R_z\cdot (R\mathbf{n}, t) = (R_z, 0)\cdot (R\mathbf{n}, t) = (R_zR\mathbf{n}, R_z t)$.
% \mgj{minimize using ``it'' and ``this.`` The text becomes vague and convoluted. Explicitly refer to something. You can Substitute X into Y.} %Take it into \cref{eq:steerability}, we have 
Replace $x$ with $(R\mathbf{n}, t)$ in \cref{eq:steerability}, and we have
\begin{equation}\label{eq:steerspec}
    \kappa(R\mathbf{n}, t) = \kappa(R_zR\mathbf{n}, R_zt), 
\end{equation}
$\forall R\mathbf{n}\in S^2{'}, \forall t\in \mathbb{R}^3{'}, \forall R_z\in \mathrm{SO}(2)'$. Notice that it implies that we need $R_zt\in \mathbb{R}^3{'}, \forall t\in \mathbb{R}^3{'}, \forall R_z\in \mathrm{SO}(2')$, so that $\kappa$ is defined on the right hand side of \cref{eq:steerspec}. In other words, the kernel points $\mathbb{R}^3{'}$ need to be closed under $\mathrm{SO}(2)'$. Obviously, having $\mathbb{R}^3{'}$ closed under $\mathrm{SO}(3)'$ is a sufficient condition for this. 

For the $S^2{'}$ dimension, since $S^2{'}$ is symmetric (closed) to $\mathrm{SO}(3)'$ by definition and $\mathrm{SO}(2)'\subset \mathrm{SO}(3)'$, we always have $R_zR\mathbf{n} \in S^2{'}$ and $\kappa$ is always defined. 

\begin{figure}
  \centering
  \includegraphics[width=\linewidth]{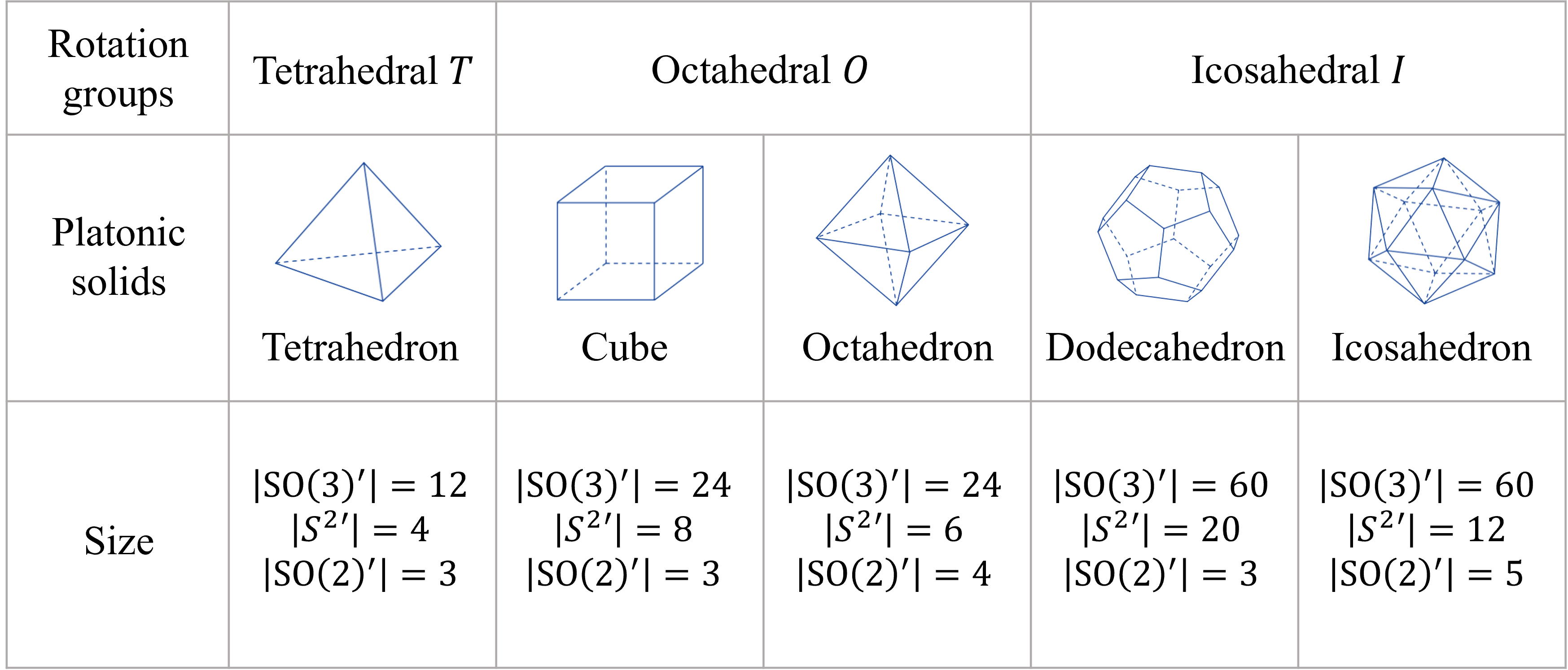}
  \caption{Illustration of all five types of Platonic solids and their corresponding finite rotation groups. }
  \label{fig:platonic}
\end{figure}

\begin{table*}
  \caption{Efficiency comparison in terms of GPU memory consumption and the computation speed between EPN~\cite{chen2021equivariant} and our method on three tasks. Two numbers are reported for \textit{training/inference} respectively. $\downarrow$ means lower is better. $\uparrow$ means higher is better. The best is shown in bold font. }
  \label{table:eff}
  \centering
%   \scriptsize
  \makebox[\textwidth][c]{
  \resizebox{\textwidth}{!}{
  \begin{tabular}{lcccccc}
\toprule
Tasks        & \multicolumn{2}{c}{ModelNet40 Pose} & \multicolumn{2}{c}{ModelNet40   Classification} & \multicolumn{2}{c}{3DMatch Keypoint Matching} \\ \midrule
Methods & Memory (GB) $\downarrow$     & Speed (fps) $\uparrow$     & Memory (GB)  $\downarrow$          & Speed (fps) $\uparrow$           & Memory (GB) $\downarrow$          & Speed (fps)  $\uparrow$        \\ \midrule
EPN \cite{chen2021equivariant}         & 22.2 / 16.9        & 1.1 / 1.6          & 13.4 / 12.7              & 1.9 / 1.5                & 37.4 / 8.5              & 0.6 / 3.1              \\
\textit{Ours} (w/o symmetric kernels)         & 4.8 / 3.7        & 5.1 / 10.1          & 4.1 / 3.2              & 7.8 / 7.8                & 7.5 / 2.8              & 2.6 / 16.7   \\
\textit{Ours} (w/ symmetric kernels)         & \textbf{4.3} / \textbf{2.8}          & \textbf{6.7} / \textbf{11.1}         & \textbf{3.9} / \textbf{2.7}               & \textbf{9.1} / \textbf{10.3}                & \textbf{6.5} / \textbf{2.4}               & \textbf{3.7} / \textbf{23.6}   \\ \bottomrule
\end{tabular}
  }
  }
  % maybe add KPConv row if have time
\end{table*}

\paragraph{Efficient feature gathering}
Another important reason for the design choice of $\mathrm{SO}(3)'$-symmetric kernels is that it enables more efficient feature gathering. Consider a spatial location $t_0\in \mathbb{R}^3$, the convolution feature output at this point is a stack of $[[\kappa*f](x_i)]_i$ with $x_i\in (S^2{'}, t_0)\subset X^\dagger$. As shown in \cref{eq:qconv}, it involves feature gathering for $\hat{f}$ at $s(x_i)\cdot x_{jk}$ for every $x_{jk}\in X'=S^2{'}\times \mathbb{R}^3{'}$. Denote an instance of $s(x_i)=(R_i, t_0), x_{jk}=(R_j\mathbf{n}, t_k)$, then $s(x_i)\cdot x_{jk}= (R_iR_j\mathbf{n}, t_0 + R_it_k)$. If $\mathbb{R}^3{'}$ is closed under $\mathrm{SO}(3)'$, then we have
\begin{equation}\label{eq:sym_ker}
    \{R_it_k|t_k\in \mathbb{R}^3{'}\}=\mathbb{R}^3{'},  \forall R_i \in \mathrm{SO}(3)', 
\end{equation}
which implies that the feature gathering for all $S^2{'}$ channels can be done once at the same set of spatial positions specified by $\mathbb{R}^3{'}$. An illustration is shown in \cref{fig:sym_kernel}. Without the symmetric kernel, the KPConv on group or quotient space looks like \cref{fig:sym_kernel}~(b), where the kernel is rotated for each $\mathrm{SO}(3)'$ (for group KPConv) or $S^2{'}$ (for quotient KPConv) channel separately to gather input features. However, with symmetric kernels, as shown in \cref{fig:sym_kernel}~(c), the rotations keep the position of kernel points unchanged up to a permutation so that the feature-gathering step is simplified. 
% \mhz{I expanded here as well. }
Specifically, the number of spatial positions needed for feature gathering without symmetric kernels is $|SO(3)'||\mathbb{R}^3{'}|$ for group KPConv or $|S^2{'}||\mathbb{R}^3{'}|$ for quotient KPConv, while the number is $|\mathbb{R}^3{'}|$ with symmetric kernels.
The convolution kernels in \cref{fig:sym_kernel} is an abstract illustration, while the actual symmetric kernel in our work is visualized in \cref{fig:kernel_spec}. 

\subsubsection{Choices of the discretization of $SO(3)$ and $S^2$}  % cannot use \mathrm in subsubsection, will take forever to compile
In \cref{sec:quo}, we mentioned that the discretization of $\mathrm{SO}(3)$ is the rotation group that respects the symmetry of a Platonic solid. There are five types of Platonic solids, corresponding to 3 finite rotation groups, as shown in \cref{fig:platonic}. Choosing $\mathrm{SO}(3)'$ to be any of them is valid, representing a discretization of $\mathrm{SO}(3)$ to different resolutions. The different Platonic solids with the same rotation group $\mathrm{SO}(3)'$ represents different discretizations of $S^2{'}$ and $\mathrm{SO}(2)'$. 

If we use a small $\mathrm{SO}(3)'$ (for example, $T$), the strategy of using $\mathbb{R}^3{'} = \{rS^2{'}\cup 0 \mid r > 0\}$ 
% \mgj{$\{rS^2{'} \cup 0\ \mid r > 0$?} 
could be problematic because the number of kernel points could be too few to learn representative features. In this case, we are free to design the kernel points differently, as long as they are \textit{closed} under $\mathrm{SO}(3)'$. For example, one may add kernel points at the center of all edges and/or faces. One may even use a combination of several polyhedrons with different radii $r$. 

In this paper, we only choose the icosahedron as the Platonic solid to conduct experiments for three reasons. First, it has the finest discretization of $\mathrm{SO}(3)$. Second, it has a smaller size of $S^2{'}$ compared with the dodecahedron, maximizing the benefit of working with quotient features. Third, it is consistent with existing methods \cite{chen2021equivariant, cesa2022program}, enabling direct comparison with the baselines. 

\subsubsection{Other aspects of the network}
We use element-wise scalar nonlinearity (ReLu and leaky ReLu) in the network. The spatial pooling is done by subsampling the input points and aggregating the features of neighboring input points to the subsampled points. Batch normalization is applied to normalize over the batch, $S^2{'}$, and $\mathbb{R}^3$ dimensions. All these choices follow the common practice of conventional CNNs (KPConv~\cite{thomas2019KPConv}) and group convolutions (EPN~\cite{chen2021equivariant}).
%More details are in the appendix. 
We also adopted the group attentive pooling in EPN to pool over the $S^2{'}$ dimension and generate $\mathrm{SO}(3)'$-invariant features. 

Depending on the specific task in the experiment, the prediction head has a slightly different design, composing the last few layers of the network. The loss functions are inherited from EPN~\cite{chen2021equivariant}, including cross-entropy loss for classifications, L2 loss for residual pose regression, and batch-hard triplet loss for keypoint matching. We refer to the appendix for more details about the prediction heads and the loss functions. 

\subsection{Relation to existing work}
% \mgj{too many sentences with ``It'' subjects. It is better to write more clearly and explicitly. Refer to the work or object you mean instead of using It or This.} 
In the context of literature on group-equivariant neural networks, our method is under the theoretical framework of \textit{equivarant CNNs on homogeneous spaces} \cite{cohen2019general, kondor2018generalization, xu2022unified}. Our work is a new form of realization when working with 3D point clouds and adapting the convolution structure of KPConv. We explore a balance of simplicity, efficiency, and expressiveness by finding the proper quotient space and discretization. Our method is an extension of group convolutions, leveraging their clean structure enabled by discretization. Our method can also be viewed as a steerable CNN with homogeneous space $S^2\times \mathbb{R}^3$ and stabilizer subgroup $\mathrm{SO}(2)$ with scalar-type features or with homogeneous space $\mathbb{R}^3$ and stabilizer subgroup $\mathrm{SO}(3)$ with $S^2$ features. The proposed work paves the way for efficiency improvement using quotient representation learning on finite groups.
% We are one of the first to explore the quotient representation of finite groups for efficiency improvement. 
% \mgj{This type of language is discouraged these days. Hard to verify who's the first in what. We can say: The proposed work paves the way for efficiency improvement in the quotient representation learning of finite groups.} 
We also emphasize the group-\textit{variant} side of equivariant models with the permutation layer to distinguish rotations, while existing work focuses on the benefit of getting group-\textit{invariant} features from equivariant models. 

\section{Experiments}\label{sec:exp}
Our major baseline to compare with is EPN~\cite{chen2021equivariant}, a state-of-the-art group convolution network, as we are similar in several ways. Both are KPConv-style convolutions with $\mathrm{SE}(3)$-equivariance. Both use the icosahedral rotation group $\mathcal{I}$ to discretize $\mathrm{SO}(3)$. However, EPN has features defined on $\mathcal{I}\times \mathbb{R}^3$, compared with $S^2{'}\times \mathbb{R}^3$ in our method. 
% while our method works with features defined on the quotient space $S^2{'}\times \mathbb{R}^3$. 

Two datasets, ModelNet40~\cite{wu20153d} and 3DMatch~\cite{zeng20173dmatch}, are used in the experiments. ModelNet40 is composed of 3D CAD models of 40 categories of objects. 3DMatch is a real-scan dataset of indoor scenes. For the ModelNet40 dataset, we conduct the classification and pose estimation tasks. For the 3DMatch dataset, we conduct the keypoint matching task. These tasks are also studied in EPN~\cite{chen2021equivariant}. 
%We will briefly introduce the experimental setup below, and EPN~\cite{chen2021equivariant} contains a good introduction of further details. 

In \cref{table:eff}, we list the GPU memory consumption and running speed of our method and EPN~\cite{chen2021equivariant} in the three tasks. The comparisons are under the same input size, number of feature channels, and number of network layers. The separable convolutions on $\mathrm{SO}(3)$ and $\mathbb{R}^3$ in EPN are together considered as one layer. 
The numbers are not comparable between training and inference because the batch size could be different (see the appendix). The specific configurations in each experiment are introduced later. All experiments are run on a single NVIDIA A40 GPU. Our network with quotient space convolutions is much smaller and runs much faster in all three tasks, 
% with full convolutions instead of separable convolutions in EPN, 
% than EPN, even given that EPN applied separable convolutions to reduce the complexity,
indicating the potential value for real applications. 
% \mhz{I added one sentence here. }Notice that our baseline EPN already applied separable convolutions to reduce the computational cost on group convolutions. 
We boost efficiency without sacrificing performance, as is shown later. 

% \mhz{I added this part.}
\textbf{Ablation study: }
We list two rows of our results in \cref{table:eff} to show the effect of efficient feature gathering enabled by the symmetric kernels. The results of \textit{without symmetric kernels} are generated using the same kernel points as \textit{with symmetric kernels}, but ignoring the fact that they are symmetric to rotations and gathering features at $|S^2{'}||\mathbb{R}^3{'}|$ locations, instead of at $|\mathbb{R}^3{'}|$ locations and permuting them. The efficient feature gathering brings further efficiency improvements, especially in terms of computational speed. 

\begin{table*}[]
\caption{ 
Experimental result of object classification on ModelNet40. 
The best is \textbf{bolded}. The best in equivariant models is \underline{underlined}. \textit{Noisy}: test using input with random translation, scaling, jittering, and dropout. \textit{Clean}: test without above processing. \textit{SO(3)}: random rotations. \textit{Id}: no rotation. \textit{ico}: random rotations in $\mathcal{I}$. ESCNN works with voxelized data, thus not having \textit{Noisy} results with point-wise augmentations. FLOPs are counted using \href{https://github.com/facebookresearch/fvcore/blob/main/docs/flop_count.md}{\texttt{fvcore}}, which does not support \href{https://docs.dgl.ai/index.html}{\texttt{DGL}} used in TFN and SE(3)-T, thus left blank. The efficiency comparison uses the same batch size as in \cref{table:eff} (see appendix). 
}
\label{tab:exp_cls}
% \centering
\resizebox{\textwidth }{!}{
\begin{tabular}{c|l|cccc|cccccc|cccc}
\hline
\multirow{5}{*}{\rotatebox[origin=c]{90}{Type}}                                                       & Column \#         & \#1   & \#2                        & \#3   & \#4                        & \#5   & \#6                        & \#7   & \#8                        & \#9         & \#10       & \#11                                                                               & \#12                                                                               & \#13                                                                               & \#14                                                                                \\ \cline{2-16} 
                                                                            & Metrics        & \multicolumn{10}{c|}{ModelNet40 classification performance metric: Acc (\%) $\uparrow$}                                                                                                                           & \multicolumn{2}{c}{Efficiency metrics}                                                                                                                                  & \multirow{4}{*}{\begin{tabular}[c]{@{}c@{}}FLOPs per\\ training \\ batch (G)\end{tabular}} & \multirow{4}{*}{\begin{tabular}[c]{@{}c@{}}Trainable \\ params \\ (M)\end{tabular}} \\ \cline{2-14}
                                                                            & Train rotation & \multicolumn{4}{c|}{SO(3)}                                              & \multicolumn{6}{c|}{Id}                                                                            & \multirow{3}{*}{\begin{tabular}[c]{@{}c@{}}Memory (GB)$\downarrow$ \\ train/test\end{tabular}} & \multirow{3}{*}{\begin{tabular}[c]{@{}c@{}}Speed (fps)$\uparrow$ \\ train/test\end{tabular}} &                                                                                    &                                                                                     \\ \cline{3-12}
                                                                            & Test rotation  & \multicolumn{2}{c|}{SO(3)}         & \multicolumn{2}{c|}{Id}            & \multicolumn{2}{c|}{SO(3)}         & \multicolumn{2}{c|}{Id}            & \multicolumn{2}{c|}{ico} &                                                                                    &                                                                                    &                                                                                    &                                                                                     \\ \cline{3-12}
                                                                            & Test condition & Noisy & \multicolumn{1}{c|}{Clean} & Noisy & \multicolumn{1}{c|}{Clean} & Noisy & \multicolumn{1}{c|}{Clean} & Noisy & \multicolumn{1}{c|}{Clean} & Noisy       & Clean      &                                                                                    &                                                                                    &                                                                                    &                                                                                     \\ \hline

\multirow{6}{*}{\rotatebox[origin=c]{90}{Non-equiv}} & KPConv \cite{thomas2019KPConv}        & 75.71 & 81.17                      & 76.88      & 82.02      & 12.66 & 12.22                      & 89.46 & 91.85                      & -           & -          & \bf{0.18}/\bf{0.25}                                                                          & 17.24/18.48                                                                        & 0.74                                                                               & 1.69                                                                                \\
                                                                            & PointNet++ \cite{qi2017pointnet++}    & 81.31 & 84.88                      & 83.03      & 85.76      & 13.44 & 13.76                      & 90.67 & 91.55                      & -           & -          & 1.03/0.5                                                                           & 7.06/7.43                                                                          & 10.59                                                                              & 1.48                                                                                \\
                                                                            & DGCNN \cite{wang2019dynamic}         & 79.86 & 84.77                      & 82.54      & 85.62      & 15.36 & 17.26                      & 91.00 & 92.18                      & -           & -          & 1.96/1.69                                                                          & \textbf{17.90}/17.50                                                                        & 32.65                                                                              & 1.81                                                                                \\
                                                                            & PT\cite{zhao2021point}            & 78.51 & 79.68                      & 78.83      & 79.68      & 16.53 & 16.77                      & 86.98 & 88.10                      & -           & -          & 4.16/5.82                                                                          & 4.80/5.03                                                                          & 220.99                                                                             & 9.58                                                                                \\
                                                                            & CurveNet\cite{Xiang_2021_ICCV}       & 84.72 & 88.33                      & 85.82      & 88.94      & 17.34 & 18.03                      & 91.53 & 92.63                      & -           & -          & 1.20/0.33                                                                          & 5.61/7.54                                                                          & 4.22                                                                               & 2.14                                                                                \\
                                                                            & PCT\cite{guo2021pct}            & 86.91 & \bf{89.10}                      & 87.64      & \bf{89.83}      & 16.33 & 17.95                      & \bf{91.69} & \bf{92.87}                      & -           & -          & 1.18/0.80                                                                          & 8.54/\bf{20.85}                                                                         & 27.41                                                                              & 2.87                                                                                \\ \hline
\multirow{6}{*}{\rotatebox[origin=c]{90}{Equivariant}}                                                & ESCNN \cite{cesa2022program}       & -     & 82.40                      & -          & 77.73      & -     & 29.03                      & -     & 88.94                      & -           & -          & 4.71/5.50                                                                          & 3.79/8.38                                                                          & 428.61                                                                             & 0.46                                                                                \\
                                                                            & TFN \cite{thomas2018tensor}          & 58.27 & 62.64                      & 58.06      & 62.64      & \underline{\bf{57.50}} & \underline{\bf{62.28}}                      & 59.20 & 62.28                      & -           & -          & 15.95/8.7                                                                          & 1.83/6.08                                                                          & -                                                                                  & 0.06                                                                                \\
                                                                            & SE(3)-T \cite{fuchs2020se}      & 60.37 & 66.29                      & 61.18      & 66.29      & 44.61 & 50.53                      & 44.81 & 50.53                      & -           & -          & 19.39/9.65                                                                         & 1.54/5.13                                                                          & -                                                                                  & 0.11                                                                                \\
                                                                            & EPN \cite{chen2021equivariant}           & 84.63 & 87.84                      & 85.34      & 88.83      & 30.99 & 32.32                      & 90.03 & 91.61                      & 90.04       & 91.60      & 13.40/12.72                                                                        & 1.86/1.49                                                                          & 58.95                                                                              & 3.06                                                                                \\ \cline{2-16}
                                                                            & \textit{Ours} (w/ GA pooling \cite{chen2021equivariant})   & 85.04 & 87.51                      & 85.49      & 87.63      & 41.46 & 44.43                      & 89.73 & 90.50                      & 90.00       & 90.50      & \underline{3.95}/\underline{2.70}                                                                          & \underline{9.21}/\underline{10.37}                                                                         & 170.68                                                                             & 2.53                                                                                \\ 
                                                                            & \textit{Ours} (w/ permutation)  & \underline{\bf{86.99}} & \underline{88.62}                      & \underline{\bf{88.21}}      & \underline{89.62}      & 39.54 & 42.78                      & \underline{90.66} & \underline{91.77}                      & \underline{\bf{90.68}}       & \underline{\bf{91.77}}      & \underline{3.95}/\underline{2.70}                                                                          & 9.09/10.28                                                                         & 170.77                                                                             & 2.65                                                                                \\ \hline
                                                                            
\end{tabular}
}
\end{table*}

\subsection{Object classification on ModelNet40}\label{sec:exp_cls}
For this task, given a point cloud of an object, the network predicts its category. The evaluation metric is classification accuracy (Acc). In this experiment, we show an extensive efficiency comparison with more existing networks, equivariant and non-equivariant. We also examine a wide combination of input conditions in training and testing to show the effect of equivariance and robustness against input imperfections. All models are trained with data augmentation, including random translation, scaling, jittering, and dropout. 

%Our network is trained with batch size 12 for 80k iterations. %This experiment is designed in ModelNet40~\cite{wu20153d} and followed by EPN~\cite{chen2021equivariant} and other existing work. 
 
Our method has outstanding performance as shown in \cref{tab:exp_cls}. In columns \#1-4, all methods are trained with rotational augmentation. Our method performs the best among listed equivariant models and is on par with PCT~\cite{guo2021pct_app} among all models. Our method is intended to work with rotational augmentations so that the equivariance gap caused by discretization can be interpolated through training. However, we still experiment with training without rotational augmentation, as shown in columns \#5-10. From columns \#9-10, we can verify the equivariance to the icosahedral rotation group of our model. Columns \#5-6 show the effect of discretization on the continuous $\mathrm{SO}(3)$ if no interpolation is trained, in which case the performance lands between non-equivariant models and continuously $\mathrm{SO}(3)$-equivariant models. The efficiency of our method outperforms all equivariant baselines and is similar to non-equivariant models. The FLOPs and number of trainable parameters are also listed for reference. We found that FLOPs do not strictly correlate with running speed, which could be due to different memory access costs and parallelism. 

\begin{table}[]
    \caption{Pose estimation on ModelNet40. Mean, median, and max angular errors are calculated over the test set. Statistics (average and standard deviation) over 10 test runs are shown to account for the randomness. 
    % \mgj{Maybe add a line to explain how initial rotations are generated and how large they are (up to 180?). It's hard to repeat the experiment with given information here.}
    }
    \label{tab:exp_rot}
    \centering
    % \footnotesize
    \makebox[\linewidth][c]{
    \resizebox{\linewidth}{!}{
    % \begin{tabular}{llll}
% \toprule
% Methods       & Mean ($\degree$) $\downarrow$  & Median ($\degree$) $\downarrow$ & Max ($\degree$)  $\downarrow$  \\ \midrule
% KPConv \cite{thomas2019KPConv} & 11.46 & 8.06   & 82.32 \\
% EPN \cite{chen2021equivariant}    & 1.25  & 1.11   & 6.63  \\
% % Ours   & \textbf{1.17}  & \textbf{1.08}   & \textbf{5.90}     \\
% \textit{Ours}   & \textbf{1.08}  & \textbf{0.96}   & \textbf{5.66}     \\ % v2 
% \bottomrule
% \end{tabular}
\begin{tabular}{l|cc|cc|cc}
\hline
Metrics            & \multicolumn{2}{c|}{Mean ($\degree$)   $\downarrow$} & \multicolumn{2}{c|}{Median ($\degree$) $\downarrow$} & \multicolumn{2}{c}{Max ($\degree$)   $\downarrow$} \\ \hline
Stats & Avg $\mu$                & SD $\sigma$                 & Avg $\mu$                 & SD $\sigma$                  & Avg $\mu$               & SD $\sigma$                \\ \hline
KPConv   \cite{thomas2019KPConv} & 13.99                        & 1.53                        & 10.70                         & 0.81                         & 115.19                      & 57.84                      \\
EPN \cite{chen2021equivariant}  & \textbf{1.10}                & 0.20                        & 1.36                          & 0.13                         & 7.06                        & 2.52                       \\
\textit{Ours}               & 1.20                         & \textbf{0.08}               & \textbf{0.96}                 & \textbf{0.05}                & \textbf{6.71}               & \textbf{1.28}              \\ \hline
\end{tabular}
    }
    }
\end{table}

% \mhz{I added a sentence here in case people wonder about the augmentation. }
% \mgj{isn't the interesting part of equivariant encoders results without data augmentation?} \mhz{I am working on it. Hopefully I can add some results without augmentation here. Initially we use augmentation because the equivariance is only to a finite subgroup of $SO(3)$, we use augmentation to train the interpolation. }

% The performance shows the ability of the networks to learn stable semantic features for each category of objects regardless of rotational variations. KPConv \cite{thomas2019KPConv} as a non-equivariant network also achieves good performance with rotational augmentations in training. However, the equivariant networks EPN~\cite{chen2021equivariant} and our method perform better, and ours performs the best in the list. Our method can learn $\mathrm{SE}(3)$-invariant and semantically-expressive features. We also implemented a steerable CNN based on ESCNN \cite{cesa2022program} as a baseline. ESCNN is a continuous-$\mathrm{SO}(3)$-equivariant CNN on voxel data with generalized Fourier analysis, which reported experiments on a similar task, the classification on ModelNet10 dataset\cite{wu20153d}. Thus we include it in the comparison.
% ESCNN underperforms EPN and our method in \cref{tab:exp_cls}. The appendix presents a more detailed discussion of the comparison with the ESCNN baseline. 

\textbf{Ablation study: }
Since this task requires $\mathrm{SE}(3)$-invariant features, there are two options in our network: either using the group-attentive pooling (GA pooling) introduced in EPN~\cite{chen2021equivariant} to pool over the $S^2{'}$ dimensions or using the permutation layer to find the canonical permutation of $S^2{'}$ dimensions. Either way, the canonical pose of objects in ModelNet40 can be used for supervision. \cref{tab:exp_cls} shows that the permutation layer yields better performance with negligible computational overhead. The reason could be that the permutation layer as in \cref{eq:permute} stacks features from $S^2{'}$ and thus preserves the information better, compared with weighted averaging over the $S^2{'}$ dimension as done in GA pooling.

\subsection{Pose Estimation on ModelNet40}
In this experiment, the network takes a pair of point clouds of an object and predicts the relative rotation between them. To avoid the pose ambiguity of objects with symmetric rotational shapes, only the airplane category is used in this experiment, with 626 models in the training set and 100 models in the test set. A point cloud is generated by randomly subsampling 1,024 points on the surface, and it is randomly rotated to form a pair.

The experimental result is shown in \cref{tab:exp_rot}. We achieved similar rotation estimation accuracy to EPN~\cite{chen2021equivariant} overall. 
While our mean error is slightly larger, the lower median error, max error, and standard deviations show that our method delivers a more reliable registration. It could imply that representing rotations as a permutation of features is more robust than representing them as a single element in the feature map. 
% It shows that reducing the domain of feature maps does not compromise the capacity of the network to capture rotational variance, as we can detect the $\mathrm{SO}(3)$ pose through the permutation layer. 
% The accuracy improvement over EPN is likely because the permutation layer generates rich features for each rotation. In \cref{sec:exp_cls} we discuss more on the benefits of the permutation layer. 
Besides, the equivariant networks outperform the non-equivariant KPConv~\cite{thomas2019KPConv} by a large margin. 

% \begin{table}[]
%     \caption{Experimental result on object classification and retrieval on the ModelNet40 dataset. }
%     \label{tab:exp_cls}
%     \centering
%     % \footnotesize
%     \makebox[\linewidth][c]{
%     \resizebox{\linewidth}{!}{
%     \input{table_cls}
%     }}
% \end{table}

\begin{table*}[t]
    \caption{Experiment result of keypoint matching on the 3DMatch dataset. The numbers are the average recall ($\%$), and the higher, the better. Notation * represents the result with the given point normal information. }
    \label{tab:exp_kpt}
    \centering
    \makebox[\textwidth][c]{
    \resizebox{\textwidth}{!}{
    \begin{tabular}{lcccccccccc}
\toprule
        & SHOT\cite{tombari2010unique} & 3DM\cite{zeng20173dmatch} & CGF\cite{khoury2017learning}  & PPFN\cite{deng2018ppfnet} & PPFF\cite{deng2018ppf} & 3DSN\cite{gojcic2019perfect} & Li\cite{li2020end}   & Li\cite{li2020end}*    & EPN\cite{chen2021equivariant}   &\textit{Ours}  \\ \midrule
Kitchen & 74.3 & 58.3    & 60.3 & 89.7   & 78.7 & 97.5   & 92.1 & \textbf{99.4}  & 99.0  & \textbf{99.4}  \\
Home 1  & 80.1 & 72.4    & 71.1 & 55.8   & 76.3 & 96.2   & 91.0 & 98.7  & \textbf{99.4}  & 98.7  \\
Home 2  & 70.7 & 61.5    & 56.7 & 59.1   & 61.5 & 93.2   & 85.6 & 94.7  & 96.2  & \textbf{96.6}  \\
Hotel 1 & 77.4 & 54.9    & 57.1 & 58.0   & 68.1 & 97.4   & 95.1 & \textbf{99.6}  & \textbf{99.6}  & 99.1  \\
Hotel 2 & 72.1 & 48.1    & 53.8 & 57.7   & 71.2 & 92.8   & 91.3 & \textbf{100.0} & 97.1  & 98.1  \\
Hotel 3 & 85.2 & 61.1    & 83.3 & 61.1   & 94.4 & 98.2   & 96.3 & \textbf{100.0} & \textbf{100.0} & \textbf{100.0} \\
Study   & 64.0 & 51.7    & 37.7 & 53.4   & 62.0 & 95.0   & 91.8 & 95.5  & \textbf{96.2}  & 95.2  \\
MIT Lab & 62.3 & 50.7    & 45.5 & 63.6   & 62.3 & \textbf{94.1}   & 84.4 & 92.2  & 93.5  & 90.9  \\ \midrule
Average & 73.3 & 57.3    & 58.2 & 62.3   & 71.8 & 95.6   & 91.0 & 97.5  & \textbf{97.6}  & 97.3  \\ \bottomrule
\end{tabular}
    }}
\end{table*}

\subsection{Keypoint matching on 3DMatch}
In this task, patches of point clouds extracted locally around keypoints in a large, dense scan are input to the network, and each is mapped to a feature vector of 64-dimension as the keypoint descriptor. Each patch has 1,024 points. Then we evaluate the average recall of keypoint correspondence across different scans through nearest neighbor search in the feature space, as proposed in PPFNet~\cite{deng2018ppfnet}. 
% Our network is trained with 16 patches in a batch for 150k iterations, consistent with EPN~\cite{chen2021equivariant}. 

% The result is in Tab. \ref{tab:exp_kpt}. 
This experiment's performance in \cref{tab:exp_kpt} indicates the capability of learning distinctive and rotation-invariant features for local patches of point clouds. Though not achieving the best in the list, our method delivers comparable performance to the top methods using only a fraction of the computational resources as EPN~\cite{chen2021equivariant} (see \cref{table:eff}). 
% \mgj{But there is no table of memory and runtime for this experiment. Can we add it?}\mhz{We do. See the last column of Fig. 1. }
We use the GA pooling layer~\cite{chen2021equivariant} in this experiment because the permutation layer requires supervision on the pose, while a canonical pose is not defined for local patches of point clouds, and GA pooling works with or without the pose supervision. However, the result shows that GA pooling over the $\abs{S^2{'}}$ features also provides distinctive features for keypoint matching. This part may be further improved by taking the information of the global scan~\cite{deng2018ppfnet} or the matching scan~\cite{huang2021predator} into consideration, in which case the permutation layer may get hints on the optimal permutation from the larger context. This topic goes beyond the focus of this paper and is left for future work. 

\section{Conclusion}\label{sec:conclusion}
This paper presents a new design of $\mathrm{SE}(3)$-equivariant point cloud convolution network, which is efficient, simple, and expressive simultaneously by working with feature maps defined on the quotient space $S^2 \times \mathbb{R}^3$ associated with the stabilizer $\mathrm{SO}(2)$. We further improve the efficiency of the convolutions by designing the kernel points to be symmetric to the discretized rotation group $\mathrm{SO}(3)'$. Moreover, we propose a permutation layer to recover $\mathrm{SO}(3)'$ information from $S^2{'}$ dimensions of the features so that the network can detect $\mathrm{SO}(3)$ rotations. Experiments show that our network delivers state-of-the-art performance in multiple tasks while consuming only a fraction of memory and computation resources as a group-convolution network with similar performance. Our method can open exciting opportunities to introduce the $\mathrm{SE}(3)$-equivariance property to mainstream point cloud networks for various tasks. 

This work also has limitations. We do not outperform EPN~\cite{chen2021equivariant} in the keypoint matching task, implying that the network, especially the permutation layer, needs improvement when dealing with inputs without a clear pose definition. Other possibilities for the network design also remain open. For example: What if we use a non-scalar type of feature on the quotient space? How to further alleviate the impact of the discretization of a continuous group? 
% However, based on what we have done, we believe that the idea presented is meaningful and the benefit is substantial. Thus we leave these questions for future work. 
From a general perspective, extending the discretized quotient space convolution strategy to other groups is also an attractive direction for future work.

%%%%%%%%% REFERENCES
% {\small
% \bibliographystyle{ieee_fullname}
% \bibliography{egbib}
% }
{\small
\balance  % for balancing left and right column
\bibliographystyle{cvpr2023/ieee_fullname}  % if put this file in the top folder
% \bibliographystyle{ieee_fullname}% if put this file in the cvpr2023 folder
% \bibliography{strings-full,ieee-full,references}  % .bib
% }
\bibliography{references}  % .bib
}

\end{document}

\title{E2PN: Efficient SE(3)-Equivariant Point Network (Appendix)}

\author{Minghan Zhu\\
University of Michigan\\
{\tt\small minghanz@umich.edu}
\and
Maani Ghaffari\\
University of Michigan\\
{\tt\small maanigj@umich.edu}
\and 
William A Clark \\
Cornell University \\
{\tt\small wac76@cornell.edu}
\and
Huei Peng\\
University of Michigan\\
{\tt\small hpeng@umich.edu}
}
\maketitle

\section{Preliminaries and notation}
% group, group action, representation, equivariance, subgroup, homogeneous space
We first review some basic concepts in group theory and representation theory briefly. They are highly relevant for understanding the big picture from a theoretical perspective of our work and general equivariant deep learning literature. 
% so that we can use the language to review the literature of group-equivariant learning and explain how our work is located among and differentiated from the existing work.\mgj{feel free to remove this paragraph if need more space.} %Detailed theoretical derivation of our method is shown later. 

\textbf{Groups:} A group $G$ is a set equipped with a binary operation $\cdot$, satisfying the following conditions: (1) the set is closed under the operation: $x\cdot y \in G, \forall x, y \in G$; (2) the operation is associative: $x\cdot (y\cdot z) = (x\cdot y) \cdot z, \forall x, y, z \in G$; (3) there is an identity element $e$ in the set such that $x\cdot e = e\cdot x = x, \forall x \in G$; (4) there is an inverse  $x^{-1}$ for each element $x$ in the set such that $x\cdot x^{-1} = x^{-1}\cdot x = e$. For example, the integer set $\mathbb{Z}$ is a group under the addition operator with identity 0 and inverse $-x$ for any $x \in \mathbb{Z}$. 
% It is not a group under the multiplication operator because for example the inverse of integer $2$ is not an integer. 
Sometimes we omit the $\cdot$ notation. 

\textbf{Group actions and representations:} We say a group \textit{acts} on a set $X$ if any element $g$ in $G$ corresponds to a transformation $\rho(g)$ on $X$, i.e., $[\rho(g)](x) \in X,  \forall x\in X$, such that $\rho(g_1)\circ \rho(g_2) = \rho(g_1g_2), \forall g_1, g_2 \in G$, where $\circ$ denotes function compositions, and $[\rho(e)](x) = x, \forall x\in X$. 
% Theoretically speaking, $\rho \in \text{Hom}(G, Aut(X))$. (\mhz{?})
When $X$ is a linear space and $\rho(g)$ is linear, we say $\rho$ is a (linear) \textit{representation} of $G$ in $X$. When $X$ is $n$ (finite)-dimensional linear space, we have a representation $\rho: G \rightarrow \text{GL}(n)$, i.e., we can write $[\rho(g)](x)$ as $\rho(g)x$, where $\rho(g)$ takes the form of $n$-by-$n$ invertible matrices and acts by multiplication on the left. For example, the representations of 2D rotations $\mathrm{SO}(2)$ in $\mathbb{R}^2$ are the 2-by-2 orthonormal matrices with determinant 1. 
% We denote $[\rho(g)](x)$ as $\rho(g)x$ as the actions used in this paper are applied on the left. 
We also use $(\rho, X)$ as a shorthand to denote the representation and space on which it acts. 

\textbf{Equivariance:}
Given spaces $V_1$ with representation $\rho_1$ of $G$ and $V_2$ with representation $\rho_2$ of $G$, we say a mapping $\phi: V_1 \rightarrow V_2$ is $G$-\textit{equivariant} if 
% $\phi(\rho_1(g)v) = \rho_2\phi(g)(v), \forall g \in G, v \in V_1$, or equivalently 
$\phi \circ \rho_1(g) = \rho_2(g) \circ \phi, \forall g\in G$. A $G$-equivariant linear map is also called an \textit{intertwiner}. The space of intertwiners is denoted $\text{Hom}_G(\rho_1, \rho_2)$, homomorphisms of group representations $\rho_1, \rho_2$ of $G$. 

\textbf{Subgroups, cosets, and quotient spaces:}
A subgroup $H$ of $G$ is a subset of $G$ that is also a group, denoted $H \leq G$. For example, $\mathrm{SO}(2)\leq \mathrm{SO}(3)$. Given $H\leq G$ and $g \in G$, we can define a (left-)\textit{coset} as $gH = \{gh | h \in H \}$. For a given $H$, all cosets are either equal or disjoint. Each coset is of the same size (contains the same number of elements), and they partition the whole group. The set of cosets forms a coset space (or \textit{quotient space}) $G/H = \{gH| g\in G\}$. 
% A quotient space is generally not a group. 
% Right-cosets are similarly defined, with an element denoted as $Hg$ and the space denoted $H\backslash G$. 
In short, a coset is both a subset in the group and an element in the quotient space. 

\textbf{Stabilizer subgroup:}
If a group $G$ acts on set $X$ by $\rho$, for $x\in X$, the stabilizer subgroup is defined as $\mathrm{Stab}_G(x)\triangleq \{g\in G| \rho(g)x=x\}$. By definition, the stabilizer subgroup $\mathrm{Stab}_G(e_{G/H})$ for the quotient space $G/H$ is $H$. 

\textbf{Homogeneous spaces:}
Assume that a group $G$ acts on a space $X$ through action $\rho$, we call $X$ a \textit{homogeneous space} of $G$ if $G$ acts \textit{transitively} on $X$, i. e., any two elements in $X$ are connected by a group action, $\forall x_1, x_2 \in X, \exists g\in G, \text{ s.t. }x_1 = \rho(g)x_2$. 
A quotient space $G/H$ is a homogeneous space of $G$. 
% \rev{For any homogeneous space $X$ and any point $x\in X$, $\mathrm{Stab}_G(x) \triangleq \{g\in G | \rho(g)x=x\}$ is called the stabilizer of $x$. $\mathrm{Stab}_G(x)$ is a subgroup of $G$ and $X \cong G/\mathrm{Stab}_G(x)$ ($\cong$ for isomorphism).}
%Any quotient space is a homogeneous space of $G$, and for any homogeneous space $X$, $\exists H\leq G$ such that $X \cong G/H$, where $H$ is also called the \textit{stabilizer}, because $H$ stabilizes the identity of $X$: $\rho(h)e_X = e_X, \forall h \in H$.

\textbf{Induced representations:}
Here is an important known result \cite{ceccherini2009induced,cohen2016steerable,cohen2018intertwiners}: given a representation $\rho$ of subgroup $H$ on vector space $V$, one can \textit{induce} a representation $\pi=\text{Ind}^G_H\rho$ of $G$ for the space of functions $\mathcal{F} = \{f: G/H \rightarrow V \}$. It provides a way to define group actions in function spaces, a foundation of the research on equivariant feature learning. 

\section{Definition of the section functions}\label{sec:sec}
In Sec. 3.2.2, we define the convolution in a homogeneous space as Eq. (4), using the section function $s: X\rightarrow G$, mapping an element in the quotient space to a group element in the corresponding coset, i.e., 
\begin{equation}\label{eq:section}
    s(x)H = x, \forall x\in X = G/H
\end{equation}
In our work, for the continuous case, $H=\mathrm{SO}(2), G=\mathrm{SO}(3), X = S^2$, and \cref{eq:section} can be rewritten as
\begin{equation}\label{eq:sec_spec}
    s(x)\mathbf{n} = x, \forall x\in S^2
\end{equation}

Since there are generally multiple group elements in a coset, section functions are not unique. Thus we need to define the section function to make the convolution well-defined. For any $R\in \mathrm{SO}(3)$, we can write $R = R_z(\alpha)R_y(\beta)R_z(\gamma)$ using Euler angles $\alpha\in [0,2\pi), \beta \in [0,\pi], \gamma\in [0,2\pi)$, and the coset it belongs to is $R\mathbf{n} = \{RR_z(\theta) | \theta \in [0, 2\pi)\} = \{R_z(\alpha)R_y(\beta)R_z(\gamma+\theta) | \theta \in [0, 2\pi)\} = R_z(\alpha)R_y(\beta)\mathbf{n}$. Thus a natural section from $S^2$ to $\mathrm{SO}(3)$ is 
\begin{equation}\label{eq:sec_ctn}
    s(R\mathbf{n}) \triangleq R_z(\alpha)R_y(\beta)
\end{equation}
  % \triangleq R'$, 
which removes the last $z$-axis rotation in $z$-$y$-$z$ Euler angle rotations. In the discretized setup, \cref{eq:sec_ctn} does not work because for $R\mathbf{n} \in S^2{'} \subset S^2$, $s(R\mathbf{n})\in \mathrm{SO}(3)$ may not be in $\mathrm{SO}(3)'$. In this case, we just arbitrarily select an element in each coset as the section so that $s': S^2{'} \rightarrow \mathrm{SO}(3)'$ satisfies \cref{eq:section}. 
% \wac{Can $\overline{s}$ be chosen to be the closest possible to $s$?}\mhz{We can, but in the implementation I did not specifically choose $s$ this way. } 
While the selection is arbitrary, it should be fixed once selected so that the behavior of the function is deterministic and consistent for different inputs.  

\section{The derivation of our proposed convolution}
\subsection{The equivariance of our convolution}\label{sec:eq_conv}
The derivation of the convolution in this paper is mostly built upon \cite{cohen2018intertwiners}. We do not discover new theorems. Our result is an application of the existing theoretical results in a specification that is not previously discussed in the literature. Here we start from the conclusions in \cite{cohen2018intertwiners} and show how it leads to our design of convolutions. 

We first need to introduce another concept. For any $g\in G, x \in G/H$, 
\begin{equation}
    (gs(x))H = g(s(x)H) = g(x) = (s(gx))H
\end{equation}
meaning that $gs(x)$ and $s(gx)$ are in the same coset, but $gs(x)$ and $s(gx)$ are not necessarily equal. We can relate these two using a function $\text{h}: G/H \times G \rightarrow H$ as: 
\begin{equation}
    gs(x) = s(gx)\text{h}(x,g)
\end{equation}
i.e., $\text{h}(x,g) \triangleq s(gx)^{-1}gs(x)$. This function $\text{h}$ describes how the representative group element twists beyond jumping to another coset when applied with another group element, therefore heavily relying on $s$. We may denote it as $\text{h}_s$, but we will go with $\text{h}$ in the following since we already selected and fixed $s$ in Sec.~\ref{sec:sec}. 
% \wac{Is it possible to choose a $s$ such that $\text{h}$ becomes the identity?}\mhz{According to \cite{cohen2018intertwiners}, $\text{h}(x,g) = \text{h}(g)$ when the group $G=G/H \rtimes H$ is a semidirect product of $G/H$ and $H$ when they are both subgroups and $G/H$ is a normal subgroup. For example, we have $\text{SE}(3) = \mathbb{R}^3\rtimes \text{SO}(3)$, therefore $\text{h}$ function does not depend on $x$ which is the location in the Euclidean space. However, I think $\text{h}$ will always depend on $g$ since $gs(x)\neq s(gx)$ in general. }

With this $\text{h}$ function, we can write down the form of induced representations. Given a space of functions $\mathcal{F} = \{ f: G/H \rightarrow V \}$, assuming $\rho: H \rightarrow \text{GL}(V)$ a representation of subgroup $H$ in $V$, we define $\pi = \text{Ind}^G_H\rho: G \rightarrow \text{GL}(\mathcal{F})$ as:
\begin{equation}\label{eq:ind}
    [\pi(g)f](x) \triangleq \rho(\text{h}(g^{-1}x, g))f(g^{-1}x)
\end{equation}
It is shown in \cite{cohen2018intertwiners} that~\cref{eq:ind} is a valid representation. Denote $\mathcal{F}_1 = \{G/H\rightarrow V_1\}$ and $\mathcal{F}_2 = \{G/H\rightarrow V_2\}$ with representations $(\rho_1, V_1)$ and $(\rho_2, V_2)$ on $H$, any linear mapping $\mathcal{F}_1\rightarrow \mathcal{F}_2$ equivariant to the induced representations $\text{Ind}^G_H\rho_1$ and $\text{Ind}^G_H\rho_2$ can be written as a cross-correlation with a twist:
\begin{equation}\label{eq:conv}
    [\kappa * f](x) = \int_{G/H} \kappa(s(x)^{-1}y)\rho_1(\text{h}(y,s(x)^{-1}))f(y)\text{d}y
\end{equation}
where the space $\mathcal{K_C}$ of valid kernels $\kappa: G/H \rightarrow \text{Hom}(V_1, V_2)$ is 
% subject to the constraint:
% \begin{equation}\label{eq:kernel}
%     \kappa(hy) = \rho_2(h)\kappa(y)\rho_1(\text{h}(y, h)^{-1}), \forall h\in H, y\in G/H
% \end{equation}
% According to \cite{cohen2018intertwiners}, the space of such $\kappa$ (denoted $\mathcal{K_C}$) 
equivalent to the space:
\begin{multline}\label{eq:kernel_space}
    \mathcal{K_D} = \{\overline{\kappa}: H\backslash G/H \rightarrow\text{Hom}(V_1, V_2) | \\
    \overline{\kappa}(x) = \rho_2(h)\overline{\kappa}(x)\rho_1^x(h)^{-1}, \forall x\in H\backslash G/H, h\in H^{\eta(x)H} \}
\end{multline}
where $H\backslash G/H = \{HgH|g\in G\}$ is the set of double cosets in which $HgH = \{h_1 g h_2 | h_1, h_2\in H \}$ is called a double coset. $\eta: H\backslash G/H \rightarrow G$ is a section function for double cosets. $H^{\eta(x)H} = \{ h\in H | h\eta(x)H = \eta(x)H\}$ is the set of stabilizers for double coset $x\in H\backslash G/H$, and $\rho_1^x$ is the representation of $H^{\eta(x)H}$ defined as $\rho_1^x(h) = \rho_1(\eta(x)^{-1}h\eta(x))$ for $h\in H^{\eta(x)H}$. 

% The equivalence mapping (isomorphism) between $\mathcal{K_C}$ and $\mathcal{K_D}$ is given by:
% \begin{equation}
%     \overline{\kappa}(x) = \kappa(\eta(x)H), \forall x\in H\backslash G/H
% \end{equation}
% \begin{equation}
%     \kappa(y) = \rho_2(h)\overline{\kappa}(Hy)\rho_1(\text{h}(\eta(Hy)H, h))^{-1}, \forall y \in G/H
% \end{equation}
% where $y = h\eta(Hy)H$ is a decomposition of $y$. 

While the above looks a bit involved, recall that in this work, we use scalar-type features, meaning that we choose the trivial identity representation for the subgroup $H = \text{SO}(2)$, i.e., $\rho_1(h) = \text{Id}_{V_1}, \rho_2(h) = \text{Id}_{V_2} \forall h\in H$, which simplifies the equations. The induced representation $\pi = \text{Ind}^G_H\rho: G \rightarrow \text{GL}(\mathcal{F})$ is in the form:
\begin{equation}\label{eq:ind_simp}
    [\pi(g)f](x) = f(g^{-1}x), \forall g\in G, x\in G/H
\end{equation}
% \wac{So here the map $\text{h}$ is the identity?}\mhz{No, the representation $\rho$ is identity. }
The convolution in \cref{eq:conv} now looks like: 
\begin{equation}
\begin{split}
    [\kappa * f](x) & = \int_{G/H} \kappa(s(x)^{-1}y)f(y)\text{d}y \\
    & = \int_{G/H} \kappa(y)f(s(x)y)\text{d}y
\end{split}
\end{equation}
which is consistent with Eq. (4) in the main paper. 
% The constraint (Eq.~\cref{eq:kernel}) on $\kappa$ becomes: 
% \begin{equation}
%     \kappa(hy) = \kappa(y), \forall h\in H, y\in G/H
% \end{equation}
The equivalent space of kernels is:
\begin{equation}
    \mathcal{K_D} = \{\overline{\kappa}: H\backslash G/H \rightarrow\text{Hom}(V_1, V_2) \}
\end{equation}
% The equivalence mapping between the two kernel spaces are now:
% \begin{equation}
%     \overline{\kappa}(x) = \kappa(\eta(x)H), \forall x\in H\backslash G/H
% \end{equation}
% \begin{equation}
%     \kappa(y) = \overline{\kappa}(Hy), \forall y \in G/H
% \end{equation}

\subsection{The specific form of our kernel}
In this paper, we work with $G = \mathrm{SE}(3)$ and $H = \mathrm{SO}(2)$. In the following, we derive $G/H$ and $H\backslash G/H$ in this setup since they do not appear commonly in the literature. 

The group $\mathrm{SE}(3)=\mathrm{SO}(3)\ltimes\mathbb{R}^3$ is the semi-direct product of $\mathrm{SO}(3)$ and $\mathbb{R}^3$ (the latter is a normal subgroup). We can denote a group element of $\mathrm{SE}(3)$ as $(R, t)$ where $R\in \mathrm{SO}(3), t \in \mathbb{R}^3$, such that the group action $\cdot$ is defined as $(R_1, t_1)\cdot (R_2, t_2) = (R_1R_2, R_1t_2+t_1)$, and accordingly the group inverse is defined as $(R, t)^{-1} = (R^{-1}, -R^{-1}t)$. 

The group $\mathrm{SO}(3)$ can be written as a subgroup of $\mathrm{SE}(3)$ as $(R, 0)$. Using Euler angles we have $\forall R\in \mathrm{SO}(3), \exists \alpha\in[0,2\pi), \beta \in [0,\pi], \gamma \in [0,2\pi)$, such that $R=R_z(\alpha)R_y(\beta)R_z(\gamma)$, where $R_z$ represents rotation around the $z$-axis, and similarly for $R_y$. We also have $\mathrm{SO}(2)\cong \{(R_z(\gamma), 0)\in \mathrm{SE}(3)|\gamma\in[0,2\pi) \}$. 

Therefore, a left coset of $H=\mathrm{SO}(2)$ in $G=\mathrm{SE}(3)$ is $gH = \{ gh | h\in H\} = \{(R_g, t_g)\cdot(R_z(\gamma_h), 0)|\gamma_h\in [0,2\pi)\} = \{ (R_z(\alpha_g)R_y(\beta_g)R_z(\gamma_g+\gamma_h), t_g) |\gamma_h\in [0,2\pi) \}$, meaning that a left coset can be parameterized by $\alpha_g, \beta_g, t_g$. Then the set of left cosets $G/H$ is homeomorphic 
% (\mhz{not 100\% sure}) (\wac{This is good; see Theorems 21.17 and 21.18 in \cite{lee_manifolds}}) 
to the Cartesian product $S^2\times \mathbb{R}^3 = \{(R_z(\alpha)R_y(\beta)\mathbf{n}, t)|\alpha\in[0,2\pi), \beta \in [0,\pi], t\in\mathbb{R}^3 \}$, where $S^2$ is the surface of a sphere, $\mathbf{n} = t(0,0,1)$ is the unit vector pointing to the north pole. Here we abuse the notation $(x\mathbf{n},y)$ as an ordered pair in the set $S^2\times \mathbb{R}^3$. It can be understood as a point $x$ on a sphere centered at some point $y$ in $\mathbb{R}^3$. The group $G=\mathrm{SE}(3)$ acts on $G/H$ by left multiplication: $(R_g, t_g)(R\mathbf{n}, t) = (R_gR\mathbf{n}, R_gt+t_g)$.

We further investigate the double coset space $H\backslash G/H$. An element in the set can be written as $H g H = \{h_1gh_2 | h_1, h_2 \in H\} =  \{(R_z(\alpha_g+\gamma_{h_1}) R_y(\beta_g) R_z(\gamma_g+\gamma_{h_2}), R_z(\gamma_{h_1})t_g) | \gamma_{h_1}, \gamma_{h_2} \in [0,2\pi) \}$. We can use $t(x,y,z)$ to specify the coordinate of an element in $\mathbb{R}^3$, and we can always rewrite $t(x,y,z) = R_z(\gamma_t)t(r_g,0,z_g)$ where $r_g = \sqrt{x^2+y^2} \geq 0$ and $\gamma_t = \arctan2(y,x)$. Then we can rewrite 
\begin{multline}
    H g H = \{(R_z(\alpha_g+\gamma_{h_1}) R_y(\beta_g) R_z(\gamma_g+\gamma_{h_2}), \\
    R_z( \gamma_{h_1}+\gamma_t)t(r_g, 0, z_g)) | \gamma_{h_1}, \gamma_{h_2} \in [0,2\pi) \}
\end{multline}
Let us rename $\gamma_1 \triangleq \gamma_{h_1}+\gamma_t, \theta_g \triangleq \alpha_g - \gamma_t, \gamma_2 \triangleq \gamma_g + \gamma_{h_2}$, then we have 
\begin{multline}
    HgH = \\
    \{(R_z(\theta_g+\gamma_1) R_y(\beta_g) R_z(\gamma_2), 
    R_z( \gamma_1)t(r_g, 0, z_g)) | \\
    \gamma_1, \gamma_2 \in [0,2\pi) \}
\end{multline}
Now it is clear that an element in $H\backslash G/H$ can be determined by four parameters $(\theta_g, \beta_g, r_g, z_g)$, with $\theta_g \in [0,2\pi), \beta_g \in [0, \pi], r_g \geq 0, z_g \in \mathbb{R}$. We denote an element in $H\backslash G/H$ as ${}_Hg_H(\theta_g, \beta_g, r_g, z_g)$. We have $H\backslash G/H\cong S^2\times \mathbb{R}^+\times\mathbb{R}$.
% (\mhz{?}) (\wac{I think it should be $S^2\times \mathbb{R}^+\times\mathbb{R}$ as the first copy of $\mathbb{R}$ is radial. It looks good though.}). 
Geometrically, each point $(r_g, z_g)$ in the $\mathbb{R}^+\times \mathbb{R}$ plane corresponds to a circle around the $z$-axis with radius $r_g$ at height $z_g$. On the other hand, $\theta_g, \beta_g$ parameterizes a point on the sphere $S^2$. 

It follows that we can define a kernel $\overline{\kappa} \in \mathcal{K_D}=\{\overline{\kappa}: S^2\times \mathbb{R}^+\times\mathbb{R} \rightarrow \text{Hom}(V_1, V_2)\}$, and then injectively map it to $\kappa\in \mathcal{K}_{C_0} = \{\kappa: S^2\times \mathbb{R}^3 \rightarrow \text{Hom}(V_1, V_2)\}$. The fact that $S^2\times \mathbb{R}^+\times\mathbb{R} \subsetneq S^2\times \mathbb{R}^3$ implies that $\mathcal{K_D} \cong \mathcal{K_C} \subsetneq \mathcal{K}_{C_0}$. In other words, there is a certain constraint on $\mathcal{K}_{C_0}$ to form the actual set of valid equivariant kernels $\mathcal{K_C}$. As shown in \cite{cohen2018intertwiners}, the general form of the constraint can be written as:
\begin{equation}\label{eq:kernel_gh}
    \kappa(hx) = \rho_2(h)\kappa(x)\rho_1(\text{h}(x, h)^{-1}), 
\end{equation}
for $\kappa \in \mathcal{K_C}, x\in G/H, h\in H$. As discussed in \cref{sec:eq_conv}, $\rho_1$ and $\rho_2$ are both identity; thus \cref{eq:kernel_gh} becomes $\kappa(hx) = \kappa(x)$, which is equivalent to Eq. (6) in the main paper in our specific case.

% We simplify and approximate the mapping as:
% \begin{equation}
%     \kappa((R_z(\theta_g)R_y(\beta_g)\mathbf{n}, t(x, y, z))) = \overline{\kappa}({}_Hg_H(\theta_g, \beta_g,\sqrt{x^2+y^2}, z))
% \end{equation}
% which is equivalent to (6) in the main paper. The approximation here is that we disentangle the rotations in the spherical and spatial coordinates, which works well in practice according to the experimental results. 

\section{Equivariance of element-wise non-linear layers and normalization layers}
For a feature map of shape $B\times C\times N\times A$, where $B$ is the batch size, $C$ is the feature channel, $N$ is the number of spatial points in $\mathbb{R}^3$, $A$ corresponds to the spherical coordinates in $S^2{'}$, a batch normalization (BatchNorm \cite{ioffe2015batch}) is to calculate the mean and variance across the $B\times N\times A$ channels and apply a constant scaling factor and shift for each $B\times N\times A$ tensor. For instance normalization (InstanceNorm \cite{ulyanov2016instance}), one just need to change $B\times N\times A$ to $N\times A$. In either case, consider the feature map as a function $f: N\times A \rightarrow \mathbb{R}^{B\times C}$, a BatchNorm or InstanceNorm (denoted $\mathcal{N}$) behaves like an element-wise operation, i.e., 
\begin{equation}
    [\mathcal{N}\cdot f](x) = a f(x)+b = \mathcal{N} \cdot f(x), \ \ \forall x\in S^2{'} \times \mathbb{R}^3
\end{equation}
since $a, b$ are constant vectors. Here $\cdot$ denotes applying a transformation. 

Recall that our induced representation \cref{eq:ind_simp} is in a similar form of a regular representation, which is realized by a change of coordinate without modifying the function value. Such a representation $\pi$ is commutative with element-wise operations (denoted as $\mathcal{E}$):
\begin{multline}
    [(\pi(g) \circ \mathcal{E}) \cdot f](x) \\ 
    = [\pi(g) \cdot (\mathcal{E} \cdot f)](x) = [\mathcal{E} \cdot f](g^{-1}x) = \mathcal{E} \cdot f(g^{-1}x) \\
    = \mathcal{E} \cdot [\pi(g)\cdot f](x) = [\mathcal{E} \cdot (\pi(g)\cdot f)](x) \\
    = [(\mathcal{E} \circ \pi(g)) \cdot f](x), \forall g\in G
\end{multline}
Or we can say: 
\begin{equation}
    \pi(g) \circ \mathcal{E} =\mathcal{E} \circ \pi(g), \forall g\in G
\end{equation}
It shows that element-wise non-linear layers like ReLu and  normalization layers, including BatchNorm and InstanceNorm, are $G$-equivariant.

\begin{figure*}
  \centering
  \includegraphics[width=0.86\textwidth]{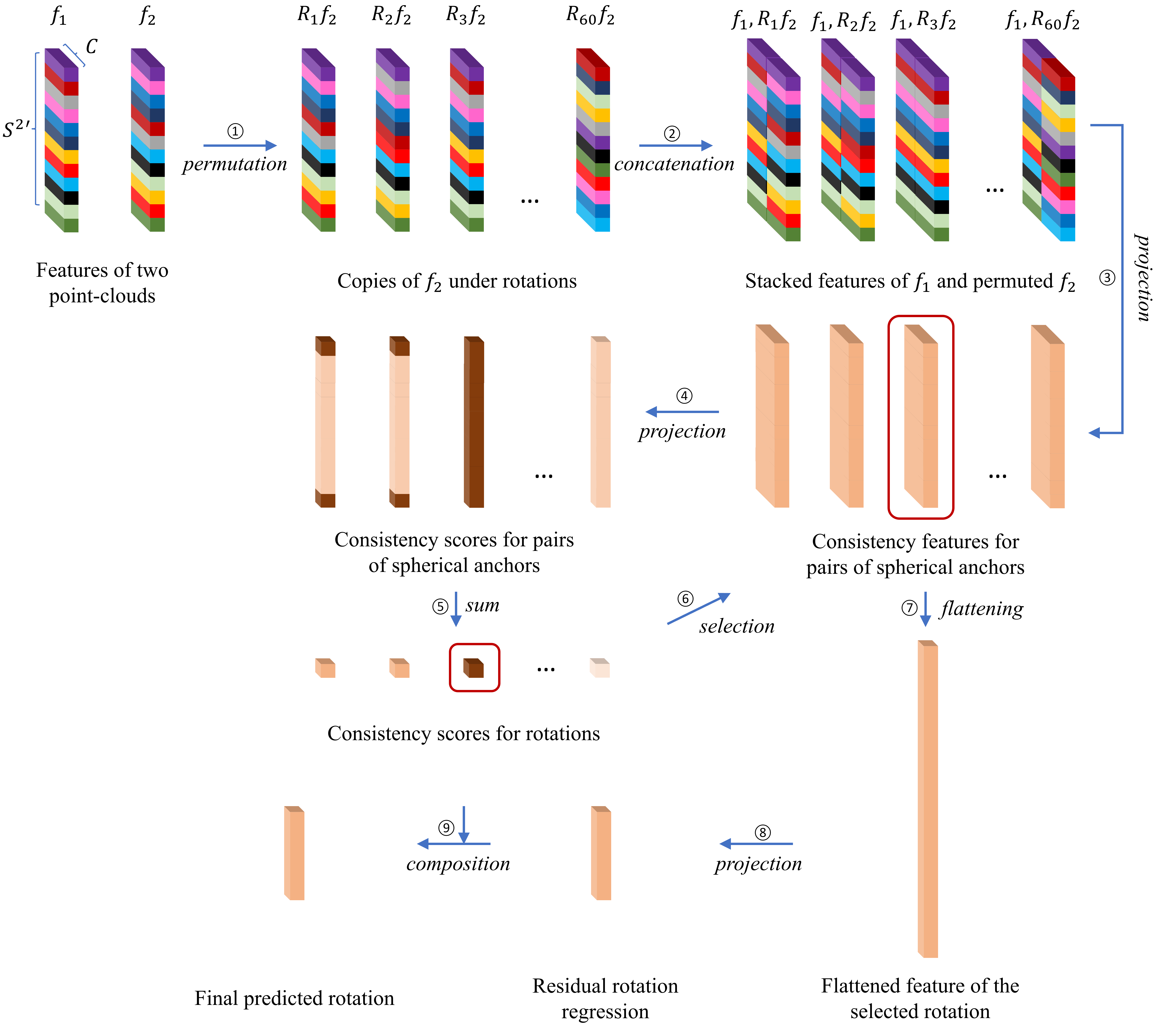}
  \caption{Illustration of the prediction head for rotation estimation. The numbers show the sequence of operations. The colors in the top row correspond to different spherical anchors. The shade of color after step 4 and after step 5 represents the matching scores for pairs of anchors. Darker means higher. }
  \label{fig:head_rot}
\end{figure*}

\section{Prediction heads and loss functions}
\subsection{Pose estimation task}
The pose (rotation) estimation task is fulfilled with a prediction head designed as shown in Fig.~\ref{fig:head_rot}. The inputs to the prediction head for each pair of point clouds are two $S^2{'}\times C$ features, where $C$ is the number of feature channels. We call the $S^2{'}$ coordinates \textit{anchors} in this section. We apply $R_i\in \mathcal{I}$ to the second point-cloud mentally, corresponding to 60 permutations of the anchors for $f_2$. If the two point clouds are different exactly by a rotation in $\mathcal{I}$, then one of the permuted $f_2$ should be exactly the same as $f_1$. We stack the features together and use several linear layers to find the match. Notice that the matching is defined as a binary classification problem for each pair of anchors instead of a multi-class classification problem for the overall feature corresponding to a certain rotation. It aligns better with the underlying geometry because a subset of anchors may align even under a wrong rotation since any rotations in $\text{SO}(2)$ keep the north-pole and south-pole vertices static. We can find the correct rotation class by summing over all anchor pairs and picking the rotation with the highest overall matching score. After finding the correct permutation (equivalent to the $R_i\in \mathcal{I}\subset \text{SO}(3)$), we flatten the feature and regress the residual rotation using quaternions in a way similar to \cite{chen2021equivariant}. 

Accordingly, the loss functions are the binary cross entropy loss for anchor-pair matching and L2 loss on the residual rotation regression. 

\begin{figure*}
  \centering
  \includegraphics[width=0.9\textwidth]{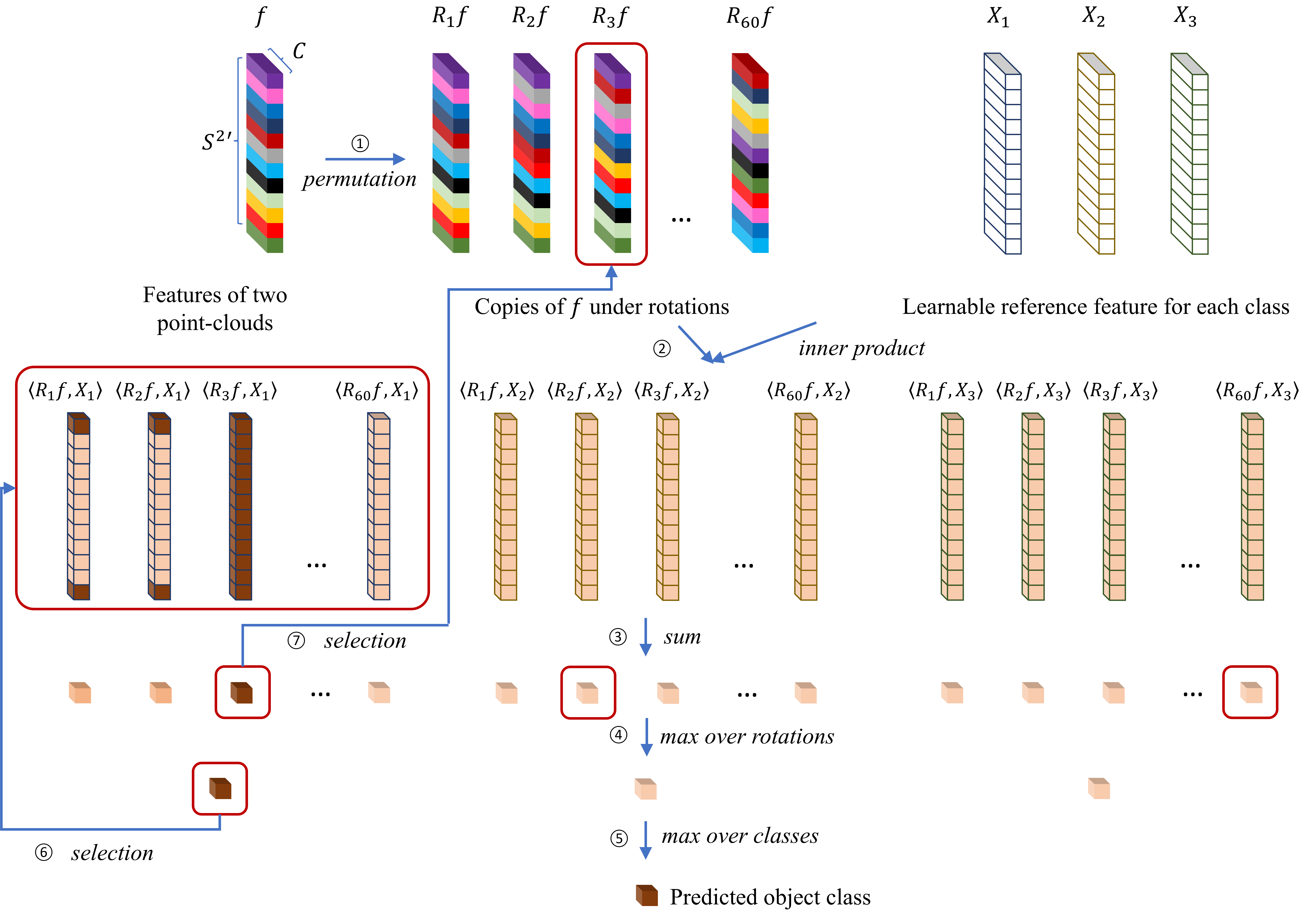}
  \caption{Illustration of the prediction head for object classification. The numbers show the sequence of operations. The solid colors in the top row correspond to different spherical anchors. The line colors on the top right represent different semantic classes. Here we only use three classes for illustration. The shade of color after step 2 represents the scores. Darker means higher. $\inprod{\cdot}{\cdot}$ denotes the inner product of finite-dim vectors. }
  \label{fig:head_cls}
\end{figure*}

\subsection{Object classification task}
For the classification task, we follow a similar philosophy as the rotation estimation task. Here we do not have a pair of inputs from which to find the relative rotation. Therefore, we imagine that there is a \textit{reference} object for each category, with a canonical permutation of the features representing the underlying canonical pose. We learn the features of the reference object in each class and use them to classify input point clouds. 

The core learnable parameter is the reference feature $X$ of shape $S^2{'}\times C \times N$ where $N$ is the number of object classes. We can denote $X_n \in \mathbb{R}^{S^2{'}\times C}$ as the reference feature for object class $n$. We directly calculate the inner product between the reference features and the permuted input features. The score of rotations under every object-class hypothesis is calculated by summing over the inner products across all anchors. The score of each class is defined as the maximum rotation score in each class hypothesis. We first generate the final prediction of the object class using the class score. Then we go back to the inner product matrix corresponding to this class and use it as the anchor-matching score prediction. Finally, the best rotation in this class is used to retrieve the specific permutation of the input feature, which forms a rotation-invariant feature for this object. 

The basic assumption here is that only the correct object class yields a high-quality matching under the actual rotation. Thus we first solve for the classification and then determine the optimal rotation only in this class, which is used for generating rotation-invariant features. 

The loss functions applied are the cross entropy loss for object classification and the binary cross entropy loss for anchor-matching prediction. 

\subsection{Keypoint matching task}
For the keypoint matching task, there is no definition of a canonical pose for a local patch of points around a keypoint. Because the feature learning in this task does not involve the corresponding patch in another point cloud, we cannot define relative poses as well. Therefore, we do not apply the permutation layer in this task. Instead, we simply follow the same design as in \cite{chen2021equivariant} using GA pooling, except that our attentive pooling is not over $\mathcal{I}$, but $S^2{'}$. 

The loss function applied here is the batch-hard triplet loss, also consistent with \cite{chen2021equivariant}. 

\begin{table*}
  \caption{The batch sizes used in the efficiency comparison in terms of the GPU memory consumption and the running speed between EPN~\cite{chen2021equivariant} and our method on three tasks as in Tab.~1. }
  \label{table:eff_spec}
  \centering
%   \scriptsize
  \makebox[\textwidth][c]{
  \resizebox{0.75\textwidth}{!}{
  \begin{tabular}{lcccccc}
\toprule
Tasks        & \multicolumn{2}{c}{ModelNet40 Pose} & \multicolumn{2}{c}{ModelNet40   Classification} & \multicolumn{2}{c}{3DMatch Keypoint Matching} \\ \midrule
Modes & Training     & Inference & Training   & Inference     & Training & Inference        \\ \midrule
Batch size         & 8        & 8          & 12              & 24                & 1$\times$16              & 8$\times$24              \\ \bottomrule         
\end{tabular}
  }
  }
  % maybe add KPConv row if have time
\end{table*}

\section{More specifications in the experiments}
The batch size used in the efficiency comparison in Tab.~1 is specified in Tab.~\ref{table:eff_spec}. For the keypoint matching task, the number of global scans processed ($n_g$) and the number of local patches extracted from each global scan ($n_l$) define the input size. We use $n_g\times n_l$ as the notation in Tab.~\ref{table:eff_spec}. 

The training optimizer and learning rate schedule follow the default setup of EPN \cite{chen2021equivariant}. The number of feature channels (i.e., width) and the number of network layers (i.e., depth) also follow the settings in EPN, except that for the object classification task on ModelNet40, we reduced the backbone width by half compared with the original EPN setting (first layer width changed from 64 to 32, later layers in the backbone changed accordingly). The network width is the same across KPConv \cite{thomas2019KPConv}, EPN, and our E2PN in our experiments, therefore their comparison remains valid. For all the three networks, the classification accuracy is not harmed by the width reduction. 

Due to the page limit in the main paper, we only mentioned "all results are trained and tested with rotational augmentation" in the classification task section. Here we provide more details. During training, random rotations are applied, following the practice in EPN. EPN and our network are equivariant to a discretization of $\mathrm{SO}(3)$; therefore, we apply rotation augmentations to let the network interpolate among the discretizations well and regress the residual rotation adding to the discretized rotations. During testing, we use fixed rotations for each test input so that results are repeatable and comparable.

We also provide more details here on the steerable CNN baseline based on ESCNN \cite{cesa2022program} in the classification task experiment. \cite{cesa2022program} established a general framework for steerable CNNs equivariant to $\mathrm{O}(3)$ and its subgroups. It is relevant to the discussion in our paper and reported results on shape classification on the ModelNet10 dataset, which is a similar task to our experiments on ModelNet 40 dataset. Therefore we include it as one of our baselines. While the library for this work is open-sourced, the specific implementation of the network for the tasks mentioned in \cite{cesa2022program} is not provided. Therefore we implemented the network using the library according to the specifics stated in \cite{cesa2022program}. We also implemented the data conversion to transform the shapes into voxel grids as described in \cite{cesa2022program}. The Gaussian kernel radius in voxel generation and the learning rate are not specified in \cite{cesa2022program}, so we did a hand tuning and reported the result under the best settings. We applied the final values $\sigma=0.03$ for the Gaussian kernel in voxel generation and $lr=10^{-3}$ as the initial learning rate. The multiplicity of irreps in the backbone is not specified either, and we found that using band-limited regular representations yields better results than using equal proportions of irreps. Thus we use multiple and-limited regular representations and align the total number of channels to the numbers specified in \cite{cesa2022program}. We implemented the $\mathrm{SO}(3)$ equivariant version with frequency up to 3. Among the invariant maps discussed in \cite{cesa2022program}, group pooling and norm pooling are implemented in their library. Group pooling is not recommended for continuous groups in \cite{cesa2022program}; therefore, we use norm pooling as the invariant layer. 

The result is shown in Table 3 in the main paper. We did not compare the efficiency because the forms of input (voxels vs. point clouds) and network structures (number of layers, channels, and connections) are both quite different. We can see that the network of \cite{cesa2022program} underperforms both EPN \cite{chen2021equivariant} and our network. One of the reasons could be that both EPN and our model for the classification task are also trained with the auxiliary task of rotation estimation (for the GA pooling layer or the permutation layer). In contrast, \cite{cesa2022program} is only trained with the classification task as described in their paper, which may cause more information loss in the invariant layer. Another reason could be that voxel inputs lose some details compared with point clouds. Transitioning from ModelNet10 to ModelNet40 dataset may require some scale-up of the network in terms of depth and width and some other careful tuning. 

%%%%%%%%%%%%%%%%%%%%%%%%%%%%%%%%%%%%%%%%%%%%%%%%%%%%%%%%%%%%
%%%%%%%%% REFERENCES
% {\small
% \bibliographystyle{ieee_fullname}
% \bibliography{egbib}
% }
{\small
\balance  % for balancing left and right column
\bibliographystyle{cvpr2023/ieee_fullname}
\bibliography{references}  % .bib
}